\documentclass[a4paper, 10pt, conference]{ieeeconf}      

\usepackage{FG2024}
\usepackage{booktabs}
\usepackage{subcaption}
\usepackage{graphicx}
\usepackage{tabularx}
\usepackage{multirow}
\usepackage{amsmath}
\usepackage{amssymb}
\usepackage{makecell}
\usepackage[pagebackref,breaklinks,colorlinks]{hyperref}

\FGfinalcopy 

\IEEEoverridecommandlockouts                              
\def\FGPaperID{144} 

\title{\LARGE \bf
ONOT: a High-Quality ICAO-compliant Synthetic Mugshot Dataset
}

\author{\parbox{16cm}{\centering
   {\large Nicolò Di Domenico, Guido Borghi, Annalisa Franco, Davide Maltoni}\\
   {
       \normalsize
       Department of Computer Science and Engineering. University of Bologna, Italy \\
       \{nicolo.didomenico, guido.borghi, annalisa.franco, davide.maltoni\}@unibo.it
   }}
   \thanks{}
}

\usepackage{fancyhdr}
\begin{document}

\ifFGfinal
\thispagestyle{empty}
\pagestyle{empty}
\else
\author{Anonymous FG2024 submission\\ Paper ID \FGPaperID \\}
\pagestyle{plain}
\fi

\twocolumn[{%
\renewcommand\twocolumn[1][]{#1}%
\maketitle

\begin{center}
    \newcommand{\imagewidth}{0.17}
    \includegraphics[width=\imagewidth\linewidth]{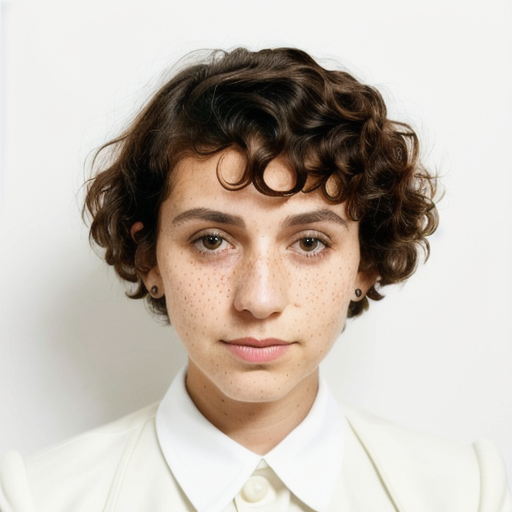}
    \medspace
    \includegraphics[width=\imagewidth\linewidth]{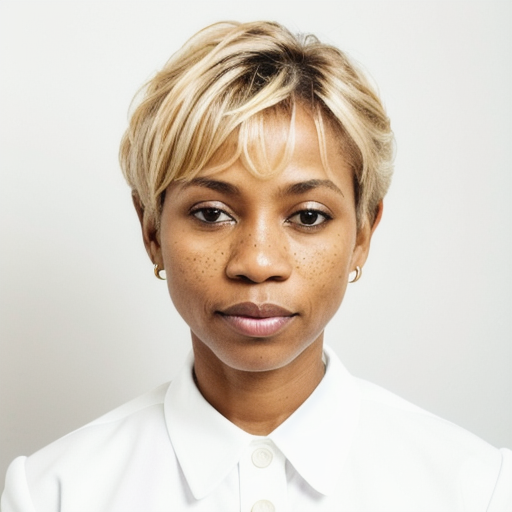}
    \medspace
    \includegraphics[width=\imagewidth\linewidth]{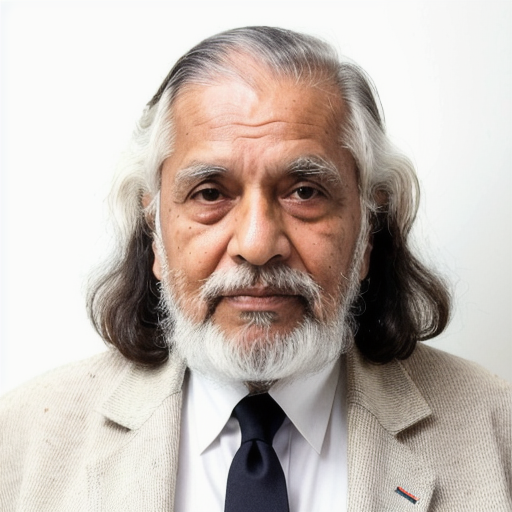}
    \medspace
    \includegraphics[width=\imagewidth\linewidth]{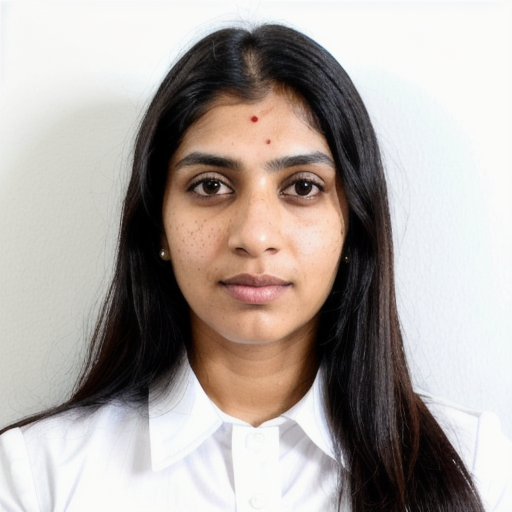}
    \medspace
    \includegraphics[width=\imagewidth\linewidth]{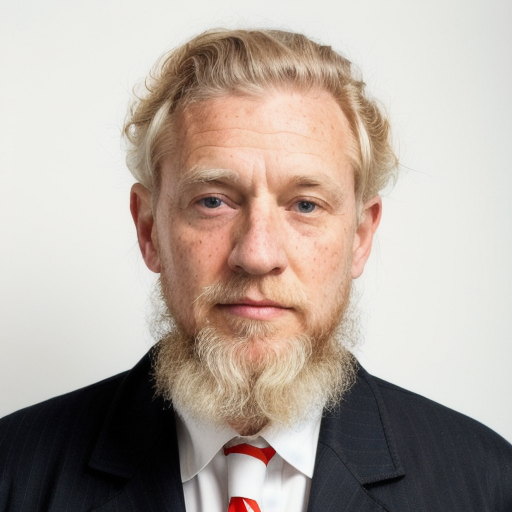}
    \\
    \vspace{0.67em}
    \includegraphics[width=\imagewidth\linewidth]{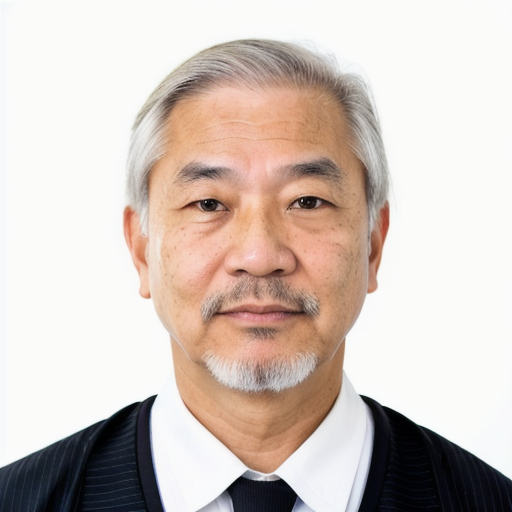}
    \medspace
    \includegraphics[width=\imagewidth\linewidth]{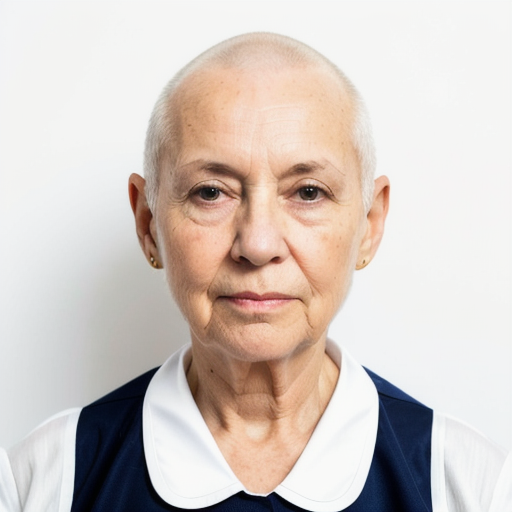}
    \medspace
    \includegraphics[width=\imagewidth\linewidth]{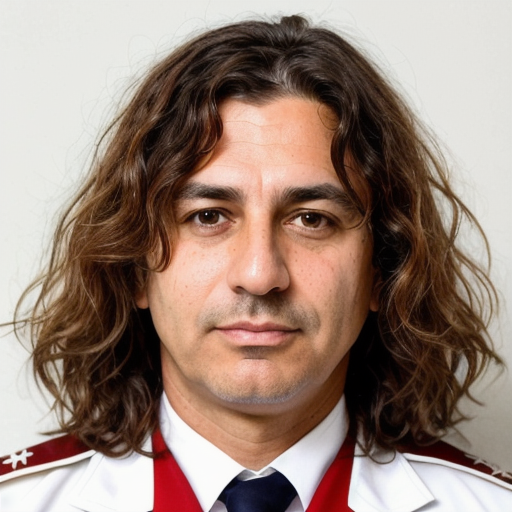}
    \medspace
    \includegraphics[width=\imagewidth\linewidth]{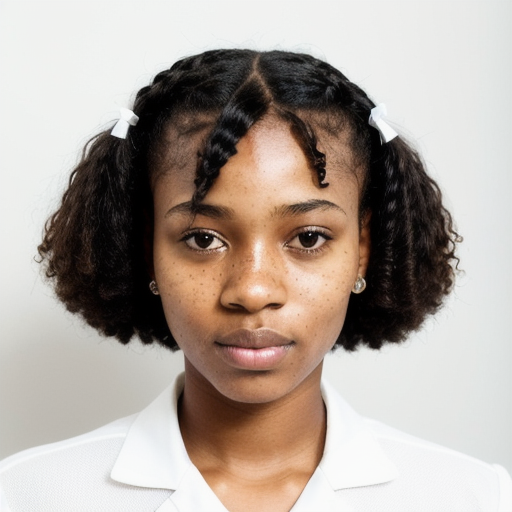}
    \medspace
    \includegraphics[width=\imagewidth\linewidth]{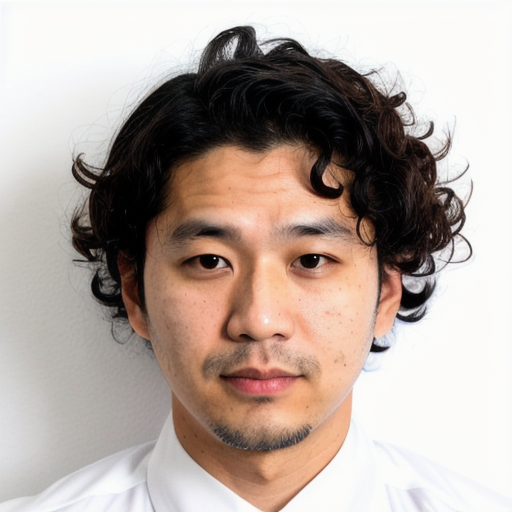}
    \captionof{figure}{Samples of the ONOT dataset compliant with the ISO/IEC 39794-5 standard and ICAO guidelines. The dataset exhibits a great inter-class variety, in terms of, among others, genders, ethnicity, age and face-specific traits.}
    \label{fig:front}
\end{center}
\vspace{1em}
}]

\begin{abstract}
Nowadays, state-of-the-art AI-based generative models represent a viable solution to overcome privacy issues and biases in the collection of datasets containing personal information, such as faces. 
Following this intuition, in this paper we introduce ONOT\footnote{\textit{One, No one and One hundred Thousand} (L. Pirandello, 1926)}, a synthetic dataset specifically focused on the generation of high-quality faces in adherence to the requirements of the ISO/IEC 39794-5 standards 
that, following the guidelines of the International Civil Aviation Organization (ICAO), defines the interchange formats of face images in electronic Machine-Readable Travel Documents (eMRTD). The strictly controlled and varied mugshot images included in ONOT are useful in research fields related to the analysis of face images in eMRTD, such as Morphing Attack Detection and Face Quality Assessment. 
The dataset is publicly released\footnote{\url{ https://miatbiolab.csr.unibo.it/icao-synthetic-dataset}}, in combination with the generation procedure details in order to improve the reproducibility and enable future extensions.
\end{abstract}


\section{INTRODUCTION}
The rapid advancement of Artificial Intelligence (AI) has introduced a new era of unprecedented opportunities and challenges. Among the various applications of AI, face-based systems have gathered significant attention due to their potential to enhance effectiveness in fields ranging from security and surveillance (\textit{e.g.}, Face Recognition~\cite{zhao2003face,melzi2024frcsyn}, Morphing Attack Detection~\cite{raja2020morphing,borghi2021double}) to human-computer interaction (\textit{e.g.}, Facial Expression Recognition~\cite{li2020deep,tian2011facial}, Facial Landmark Detection~\cite{wu2019facial,frigieri2017fast}).

However, the adoption of these technologies has raised critical concerns, particularly related to privacy infringement and inherent biases~\cite{bowyer2004face}.
For instance, algorithms for face image analysis have heavily relied on large-scale datasets~\cite{schroff2015facenet,taigman2015web,zhu2021webface260m} containing images of individuals' faces. 
While essential for training robust models, the acquisition and the release of these datasets have become increasingly problematic: the utilization of real facial images raises significant privacy concerns related, among others, to unauthorized face recognition, compromising privacy and personal security. 

In this scenario, synthetic data generated through novel generative methods emerges as a promising solution to address these pressing issues~\cite{abay2019privacy}.
Indeed, synthetic facial data offers a way to mitigate these privacy risks: by employing generative methods such as Generative Adversarial Networks (GANs)~\cite{goodfellow2020generative}, Variational Autoencoders (VAEs)~\cite{kingma2019introduction} and Diffusion Models~\cite{Rombach2022CVPR}, it is possible to generate highly realistic facial images that do not directly correspond to any real individual's identity, thus granting anonymity.

Furthermore, the creation of traditional face-based datasets has perpetuated biases that exist in society~\cite{karkkainen2021fairface}. Biases related to ethnicity, gender, age, and other demographic factors have been - inadvertently or not - embedded in these datasets, leading to biased AI models~\cite{ntoutsi2020bias}. 
Synthetic data presents an opportunity to counteract these biases: by carefully controlling the attributes of synthetic faces, it is possible to contrast the underrepresentation of specific groups, ultimately leading to fairer face recognition technologies.

Therefore, in this paper, we introduce ONOT, a novel dataset of synthetic faces, meticulously crafted in adherence to the principles outlined in the ISO/IEC 19794-5 standard~\cite{ISO-19794-5}, successively modified by ISO/IEC 39794-5~\cite{ISO}, \textit{i.e.} the reference standard in the context of face verification in electronic Machine-Readable Travel Documents (eMRTD).
The standard describes the specific requirements for enrollment images imposing strict quality criteria to be fulfilled to enable effective automatic face verification: a summary of these principles is reported in Table \ref{tab:test_iso}.
This ISO standard has been designed starting from the guidelines initially provided by the International Civil Aviation Organization (ICAO) for passport photographs~\cite{wolf2018icao} (in the following, this standard is referred also as ISO/ICAO).


More specifically, we take this standard as the inspiring principle for our generation process, which is aimed at the creation of high-quality and well-controlled images with specific characteristics including, among others, frontal face pose with uniform background and illumination, neutral expression, and the absence of shadows (see Fig. \ref{fig:front}). 
Then, we aim to create a synthetic dataset combining AI-based generative procedures, in terms of facial likeness and realism, together with the strict ISO/ICAO requirements.
These unique features enable the use of the ONOT dataset for a variety of vision-based tasks related to the analysis of identity documents or, in general, in which there is the need for high-quality and standard frontal images, including the development of methods for Morphing Attack Detection~\cite{venkatesh2021face} or Face Quality Assessment~\cite{franco2022face}, for which ad-hoc public synthetic datasets are generally not available.

    

Summarizing, the ONOT dataset offers several key advantages and features:
\begin{itemize}
    \item \textbf{ISO/ICAO compliance}: the dataset is a pioneering example of synthetic data specifically designed to meet ISO/ICAO standard requirements, and its compliance has been validated using a commercial SDK. To the best of our knowledge, this is the first synthetic dataset of its category in the literature.
    \item \textbf{Facial realism}: ONOT dataset presents high quality and realism in the generated faces, thanks to the use of a state-of-the-art generative method. The dataset comprises a collection of several subjects, including for each at least one ISO/ICAO compliant image and multiple additional samples. Each facial attribute is provided in dataset annotation.
    \item \textbf{Identity check}: rigorous verification procedures ensure both intra-subject consistency (all images of the same subject share the same identity) and inter-subject consistency (each subject presents a distinct and unique identity with respect to all the other subjects). 
    \item \textbf{Reproducibility}: this dataset is highly reproducible and expandable, as it provides comprehensive documentation regarding the model, and the image generation and selection procedures. Indeed, we release the prompts used for each generation, fostering transparency and encouraging further research and data contributions. 
\end{itemize}

\begin{table}[th!]
    \centering
    \begin{tabular}{c|l}
    \toprule
    \textbf{No} & \textbf{Description of the test} \\
    \midrule
    1 & Unique and valid face \\
    2 & Face fully included in image frame \\
    \midrule
    \multicolumn{2}{l}{\textbf{Geometric tests}} \\
    \midrule
    3 & Eye distance \\
    4 & Horizontal/vertical position \\
    5 & Head image width/height ratio  \\
    \midrule
    \multicolumn{2}{l}{\textbf{Photographic tests}} \\
    \midrule
    6 & Face is correctly focused \\
    7 & Sharpness of the image \\
    8 & Face saturation \\
    9 & Image color conformance \\
    10 & Shadows over the face \\
    11 & Glasses with dark colored lenses or glare \\
    12 & Cluttered background \\
    \midrule
    \multicolumn{2}{l}{\textbf{Pose and facial attributes tests}} \\
    \midrule
    13 & Gaze direction \\
    14 & Mouth expression \\
    15 & Correct position of shoulders \\
    16 & Both eyes visible and open \\
    17 & Eyes color \\
    18 & Eyes occluded by glasses or hair \\
    19 & Presence of glasses \\
    20 & Glasses' frames too heavy \\
    21 & Presence of hat/cap on head \\
    \bottomrule
    \end{tabular}
    \caption{Tests carried out by the commercial ICAO SDK to decide whether an image is ISO/ICAO compliant. 
    As reported, tests range from geometric and photographic to pose- and attribute-related aspects.
    }
    \label{tab:test_iso}
\end{table}

\begin{figure*}[th!]
    \centering
    \includegraphics[width=1\linewidth]{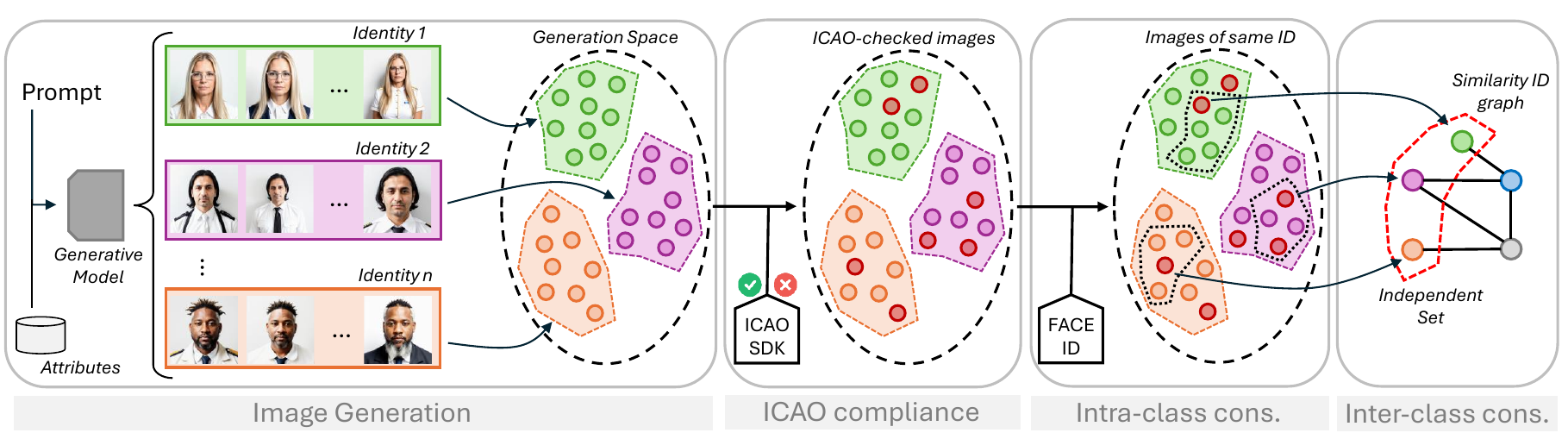}
    \caption{Steps for the generation of the ONOT dataset. Starting from the initial image generation procedure, we apply a commercial SDK to verify if the generated images are compliant with the ISO/ICAO standard. The following steps regard the verification of the intra-class consistency, \textit{i.e.} all images of the same subject share the same identity and the inter-class consistency, \textit{i.e.}  each subject presents a unique identity with respect to all the other generated subjects.}
    \label{fig:overview}
\end{figure*}

\section{RELATED WORK}

Due to the spread and efficacy of new AI-based generative algorithms, several datasets that contain synthetic faces are available in the literature~\cite{boutros2023synthetic}. 
The large majority of face-based synthetic datasets have been collected specifically for the Face Recognition task, and then they are created to have a high number of images, identities, head poses, neglecting specific standard requirements. 

SynFace~\cite{qiu2021synface} addresses the challenges in collecting large-scale real-world training data for face recognition, especially considering label noise and privacy issues. 
The work identifies the performance gap between face recognition models trained with synthetic and real face images as poor intra-class variations and the domain gap between synthetic and real images. 
The synthesis is based on the DiscoFaceGAN~\cite{deng2020disentangled} model and regards mostly frontal-view images, but the identity preservation between subjects is not evaluated.

In~\cite{bae2023digiface} DigiFace-1M, a large-scale synthetic dataset, is presented. The dataset is designed to address the scarcity, the biases and the label noise of diverse datasets for training face recognition models. DigiFace-1M, created through the framework presented in~\cite{wood2021fake}, provides a comprehensive set of facial images with varied attributes, including ethnicity, age, and facial expressions. 
The dataset is particularly notable for its scale (1 million images), but unfortunately, the level of realism seems to be limited.

The USynthFace dataset~\cite{boutros2023unsupervised}, generated through DiscoFaceGAN~\cite{deng2020disentangled}, includes synthetic face images with variability in identities, poses, illuminations, and expressions. In the paper, the authors particularly emphasize the use of synthetic data for training models in an unsupervised manner.

Differently, SFace~\cite{boutros2022sface} is created using a StyleGAN2-ADA~\cite{karras2020training} generative model under class-conditional settings, which generated $634$k synthetic images, equally distributed across $10$k classes. The main limitations are due to the limited variation in the same class and the demographic bias inherited from StyleGAN2. These limitations have been specifically addressed in~\cite{melzi2023gandiffface} with GANDiffFace framework, based on a combined use of GAN and diffusion models.

Recently, a variety of synthetic datasets have been generated through the use of diffusion models~\cite{nichol2021improved,dhariwal2021diffusion}: these datasets, in particular, are focused on identity preservation and diversification through inversion of pre-trained face recognition models (ID3PM~\cite{kansy2023controllable}), style variation combined with subject consistency (DCFace~\cite{boutros2023idiff}) and the use of authentic embeddings obtained from the authentic training datasets to enhance the realism of generated images (IDiff-Face~\cite{boutros2023idiff}).
\begin{table*}[th!]
\begin{tabularx}{\textwidth}{lX} \\ 
\multirow{5}{*}{\makecell{\textbf{Negative} \\ \textbf{prompt}}} & (deformed iris, deformed pupils, semi-realistic, CGI, 3D, render, sketch, cartoon, drawing, anime:1.4), text, close up, cropped, out of frame, worst quality, low quality, jpeg artifacts, ugly, duplicate, morbid, mutilated, extra fingers, mutated hands, poorly drawn hands, poorly drawn face, mutation, deformed, blurry, dehydrated, bad anatomy, bad proportions, extra limbs, cloned face, disfigured, gross proportions, malformed limbs, missing arms, missing legs, extra arms, extra legs, fused fingers, too many fingers, long neck, hair in front of the eyes, hat, (shadows), (three-quarter pose), (face in profile:1.1) \\
\midrule
\multirow{3}{*}{\makecell{\textbf{Prompt} \\ \textbf{template}}} & RAW front photo, face portrait photo of (\{years\} years old:1.1), \{ethnicity\} (\{gender\}:1.1), \{hair color\}  hair, (\{hair style\} hair style:1.1), (\{traits\}:1.1), neutral expression, wearing dress, (white background:1.4), head horizontally aligned, (uniform lighting:1.4), top of the hair visible, (passport photo:1.1) \\
\midrule
\multirow{3}{*}{\textbf{Prompt}} & RAW front photo, face portrait photo of (81 years old:1.1), African (female:1.1), black wavy hair, (braids hair style:1.1), (glasses and freckles:1.1), neutral expression, wearing dress, (white background:1.4), head horizontally aligned, (uniform lighting:1.4), top of the hair visible, (passport photo:1.1) \\
\end{tabularx}
\caption{The negative prompt used for generating the images, the template of the positive prompt and one example of prompt of a subject, as detailed in Section \ref{sec:image_generation}. The extensive negative prompt ensures that the images have a natural look, with realistic facial traits. The words within parentheses are assigned a greater weight by the model.}
\label{tab:prompt}
\end{table*}
In the context of Face Morphing, a synthetic dataset is proposed in~\cite{damer2022privacy}.
The SMDD dataset contains $30$k morphing attack and $50$ bona fide samples. 
The morphing attack detection models~\cite{raja2020morphing} trained on SMDD demonstrated high performance even when tested against unknown attack types and morphing techniques, indicating its robustness and generalizability. Unfortunately, we found that these images do not pass the ISO standard checks (see Table~\ref{tab:test_iso}), resulting often in morphed or bona fide images with low-quality or visible artifacts.   

In summary, the literature demonstrates significant progress in the development and utilization of synthetic facial image datasets. 
On the one hand, these datasets are increasingly being recognized as valuable tools for addressing the challenges of privacy, biases, and data availability in face recognition research.
On the other hand, these datasets, often explicitly created only for the face recognition task in uncontrolled scenarios, tend to
disregard the consistency of synthesized identities and the need for standard images that are used in document-related tasks.


\section{DATASET GENERATION}\label{sec:construction}

A representation of the generation is provided in Figure~\ref{fig:overview}.

\subsection{Image generation} \label{sec:image_generation}
In this step, the goal is to generate facial images with a high level of realism and quality, 
compliant with the requirements of the ISO/ICAO standard. 

To start the image generation process, $15$k initial identities, here referred to as pseudo-classes (since, at this step, it is not guaranteed that different generated images correspond to different real identities), are defined through a random seed that, among others, contains information about the identity. 
Each pseudo-class is defined by the combination of a prompt and the initial seed. 
For each pseudo-class we generate $64$ images, using a fixed negative prompt, a random positive prompt and increasing the initial seed by $1$.
The generation is based on a fine-tuned version of Stable~Diffusion~1.5~\cite{Rombach2022CVPR}, namely Realistic~Vision~5.1.
The model is served using Stable Diffusion Web UI~\cite{AUTOMATIC1111_Stable_Diffusion_Web_2022}.
Each image has a resolution of $512 \times 512$ and is generated using the DPM++ SDE Karras sampler~\cite{lu2022dpm,karras2022elucidating} with $25$ steps.
To generate the $15$k identities (64 images per identity, for a total of $960$k generated images), we employ $32\times \text{A}100$ 64GB Nvidia GPUs for $14$ hours in total.

Positive prompts are generated by randomizing values inserted into a predetermined template.
Specifically, to emulate the characteristics of an official eMRTD picture, 
we engineer the prompt to obtain images able to pass the tests listed in Table~\ref{tab:test_iso} reflecting the ISO/ICAO requirements. 
The main aspects we explicitly control are related to the neutral expression upright frontal pose, bright background and uniform lighting. 
These desired attributes are assigned a higher weight than the rest of the prompt, given their importance in the context of this dataset, as detailed in the prompt template reported in Table~\ref{tab:prompt} (in which the negative prompt and two samples of positive prompts are also reported).
Properties such as gender and face traits are chosen following a weighted selection algorithm, and the probabilities are set as follows: $48$\% for male/female, $4$\% for non-binary; $23$\% for moles, freckles, moles and freckles; $2$\% for the other combinations of facial traits and attributes, included the presence of the glasses.
A comprehensive list of these properties is given in Table~\ref{tab:file-naming}, which also includes the file naming convention used for the dataset.
To further improve the variability of the dataset, we also include details in the prompt about the hair color (\textit{e.g.} blonde, brown) and style (\textit{e.g.} curly, straight, bold, with fringe), glasses type (\textit{e.g.} round lenses, metal glasses) and gender-specific traits (\textit{e.g.} beard), sampled through a uniform probability distribution.

\subsection{ISO/IEC 39794-5 compliance} \label{sec:iso_check}
In this step, we aim to verify if each generated image fulfills the ISO/ICAO quality requirements. 
This ISO/IEC 19794-5 standard~\cite{ISO-19794-5}, recently modified by ISO/IEC 39794-5~\cite{ISO}, has been introduced to establish uniform guidelines and specifications for the exchange of biometric data, specifically facial images, between different systems and organizations. It was developed to address the need for interoperability and consistency in the field of biometrics, especially in applications based on identity verification and authentication~\cite{schlett2022face}. Then, the standard promotes compatibility between various biometric systems and helps prevent data inconsistencies and errors when using automated facial recognition technology~\cite{busch2023standards}.

The compliance verification procedure is carried out through a commercial SDK\footnote{\url{https://www.correlance.com/cms/en/home}}.
Specifically, this SDK verifies the presence of scene constraints (such as pose, and expression), photographic properties (\textit{e.g.} lighting, positioning, and camera focus), as well as digital image attributes (\textit{e.g.} image resolution, and image size). A comprehensive list of tested features is reported in Table~\ref{tab:test_iso}.
Upon the completion of the validation, pseudo-classes that do not contain at least one ISO/ICAO-compliant image are discarded.

\begin{table}[th!]
\centering
    \begin{tabular}{cll}
    \toprule
    \textbf{Field} & \textbf{Description} & \textbf{Values} \\
    \midrule
    m & Model name & S - Stable Diffusion \\
    x\{8\} & Seed & 00000000 - 99999999 \\
    \midrule
    Gg & Gender & GM - male \\
    & & GF - female \\
    & & GN - non-binary \\
    \midrule
    Aaa & Age & A18-A99 years \\
    \midrule
    Eee & Ethnicity & EEA - European/American \\
    & & EAF - African \\
    & & EIA - Indian-Asian \\
    & & EAS - East-Asian \\
    & & EME - Middle Eastern \\
    \midrule
    Ttt & Face traits & T00 - none \\
    & & T01 - moles  \\
    & & T02 - scars  \\
    & & T03 - freckles \\
    & & T10 - glasses \\
    & & T11 - glasses and moles \\
    & & T12 - glasses and scars \\
    & & T13 - glasses and freckles \\
    & & T14 - moles and freckles \\
    & & T15 - glasses, moles, \\
    &&and freckles \\
    & & T25 - freckles and scars \\
    \midrule
    Innnn & Image number & 0001 - 9999 \\
    \midrule
    Fff & Image format & F00 - digital \\
    & & F01 - Print\&Scan (P\&S)\\
    \bottomrule
    \end{tabular}
    \caption{The file naming scheme used to save the images to disk, which allows to understand the variety of the generated elements included in the ONOT datasets.
    }
    \label{tab:file-naming}
\end{table}

\begin{figure*}[th!]
    \newcommand{\imagewidth}{0.16}
    \centering
    \begin{subfigure}[b]{\imagewidth\linewidth}
        \centering
        \includegraphics[width=\linewidth]{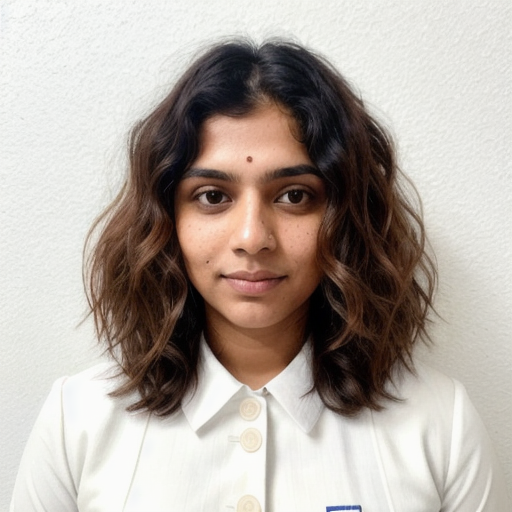}
    \end{subfigure}
    \medspace
    \begin{subfigure}[b]{\imagewidth\linewidth}
        \centering
        \includegraphics[width=\linewidth]{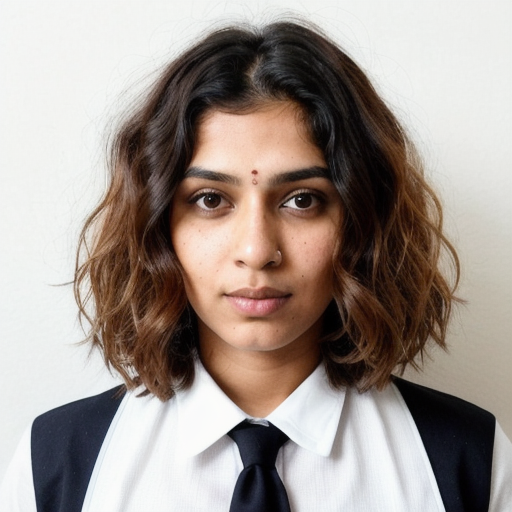}
    \end{subfigure}
    \medspace
    \begin{subfigure}[b]{\imagewidth\linewidth}
        \centering
        \includegraphics[width=\linewidth]{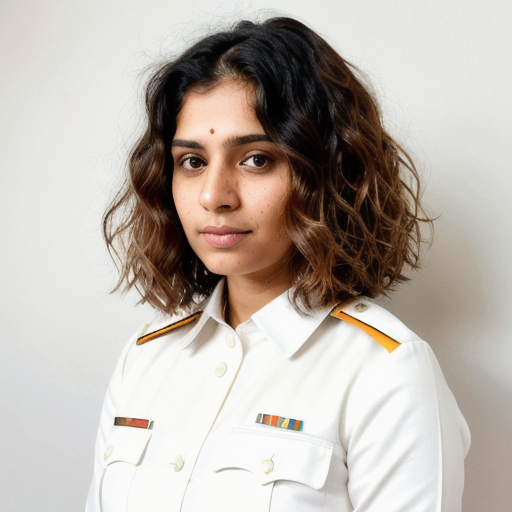}
    \end{subfigure}
    \medspace
    \begin{subfigure}[b]{\imagewidth\linewidth}
        \centering
        \includegraphics[width=\linewidth]{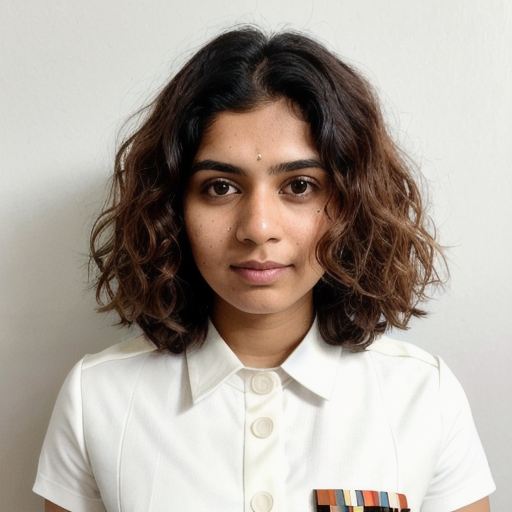}
    \end{subfigure}
    \medspace
    \begin{subfigure}[b]{\imagewidth\linewidth}
        \centering
        \includegraphics[width=\linewidth]{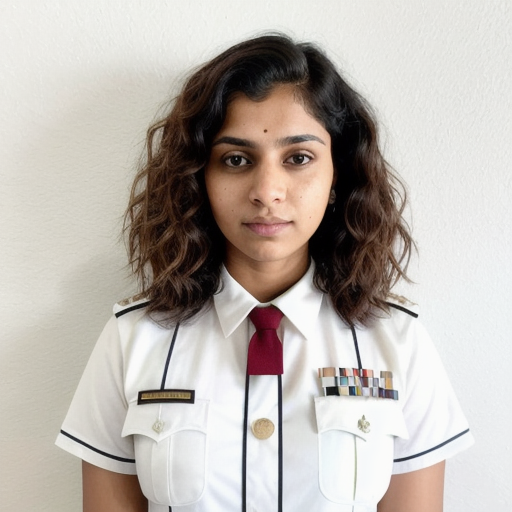}
    \end{subfigure}
    \\
    \vspace{0.67em}
    \begin{subfigure}[b]{\imagewidth\linewidth}
        \centering
        \includegraphics[width=\linewidth]{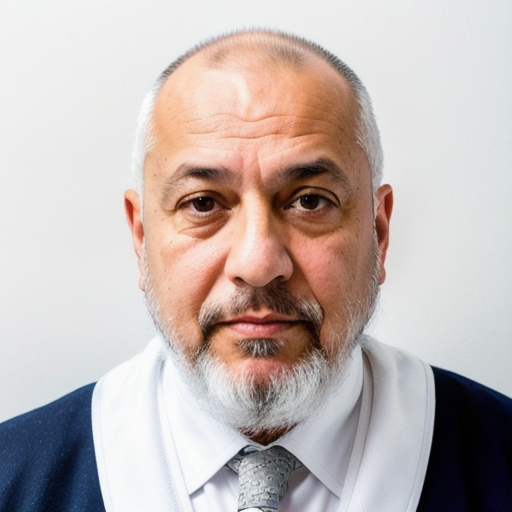}
    \end{subfigure}
    \medspace
    \begin{subfigure}[b]{\imagewidth\linewidth}
        \centering
        \includegraphics[width=\linewidth]{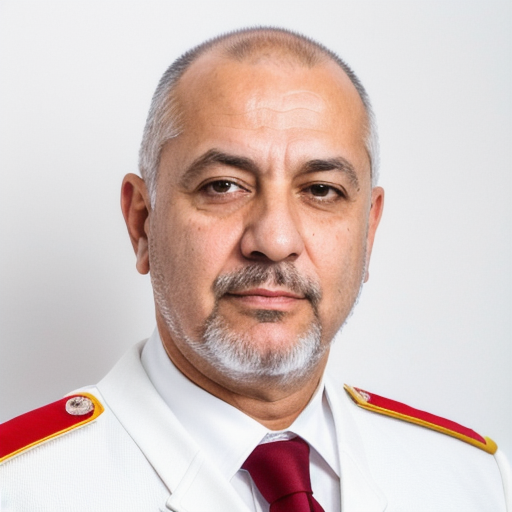}
    \end{subfigure}
    \medspace
    \begin{subfigure}[b]{\imagewidth\linewidth}
        \centering
        \includegraphics[width=\linewidth]{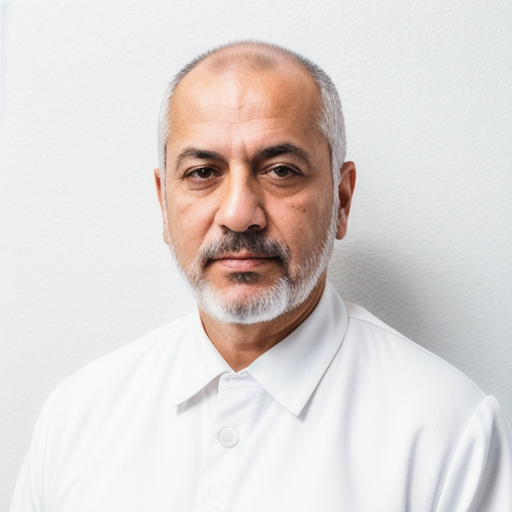}
    \end{subfigure}
    \medspace
    \begin{subfigure}[b]{\imagewidth\linewidth}
        \centering
        \includegraphics[width=\linewidth]{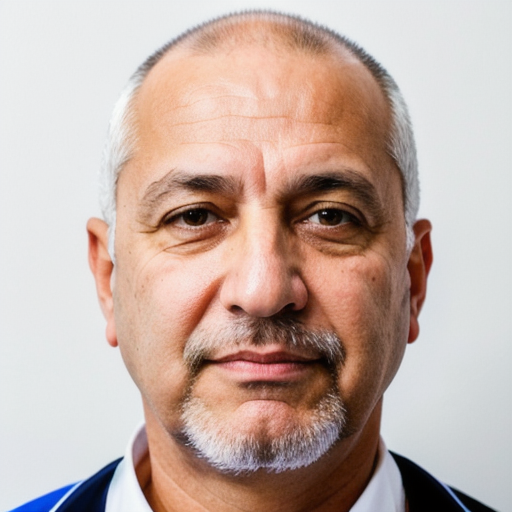}
    \end{subfigure}
    \medspace
    \begin{subfigure}[b]{\imagewidth\linewidth}
        \centering
        \includegraphics[width=\linewidth]{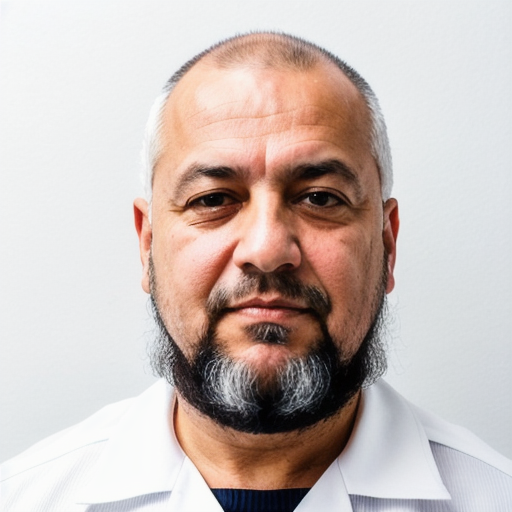}
    \end{subfigure}
    \\
    \vspace{0.67em}
    \begin{subfigure}[b]{\imagewidth\linewidth}
        \centering
        \includegraphics[width=\linewidth]{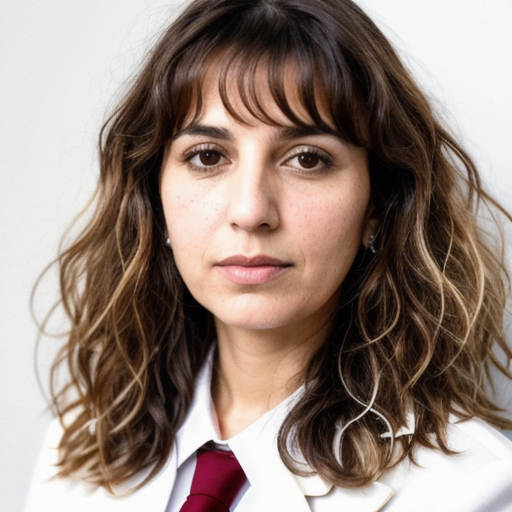}
    \end{subfigure}
    \medspace
    \begin{subfigure}[b]{\imagewidth\linewidth}
        \centering
        \includegraphics[width=\linewidth]{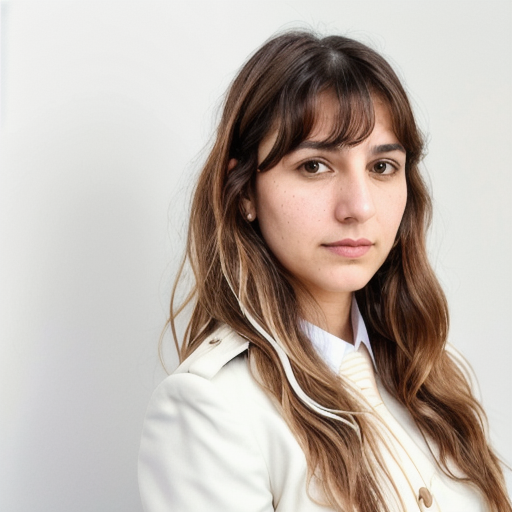}
    \end{subfigure}
    \medspace
    \begin{subfigure}[b]{\imagewidth\linewidth}
        \centering
        \includegraphics[width=\linewidth]{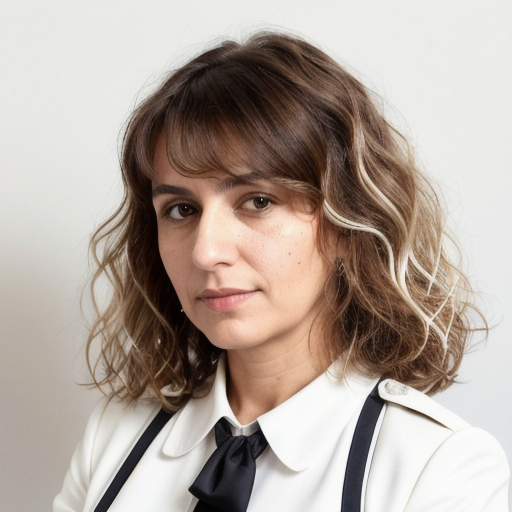}
    \end{subfigure}
    \medspace
    \begin{subfigure}[b]{\imagewidth\linewidth}
        \centering
        \includegraphics[width=\linewidth]{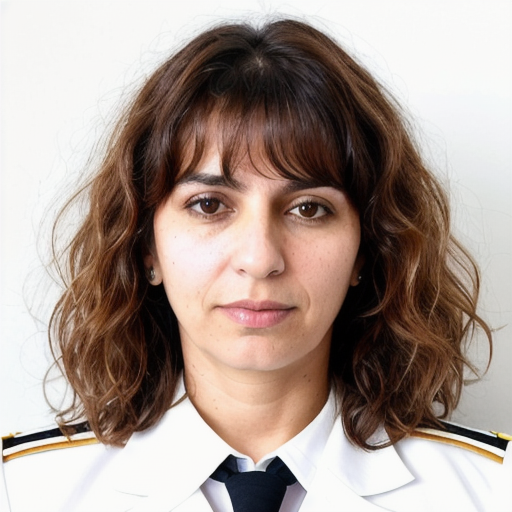}
    \end{subfigure}
    \medspace
    \begin{subfigure}[b]{\imagewidth\linewidth}
        \centering
        \includegraphics[width=\linewidth]{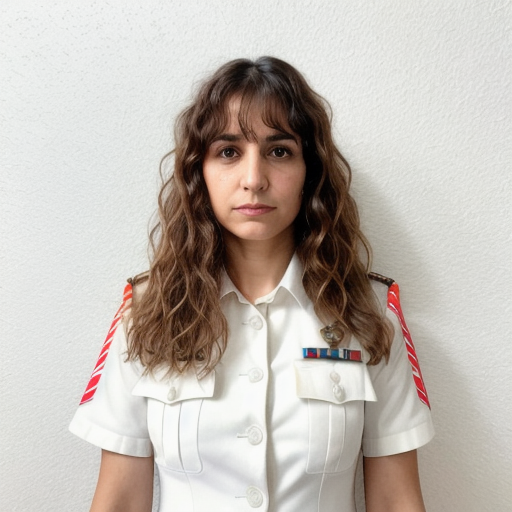}
    \end{subfigure}
    \caption{In addition to ISO/ICAO-compliant samples, other images are generated for each identity. As shown, intra-class variance is present in terms of different head and body poses and facial traits. }
    \label{fig:samples_same_subject}
\end{figure*}

\subsection{Intra-class consistency}
In this step, we aim to verify if the $64$ images grouped in each pseudo-class belong or not to the same identity.
Therefore, a face recognition pipeline is applied to each image. In particular, we detect faces employing the MTCNN~\cite{zhang2016joint} face detector and align them following the SphereFace~\cite{liu2017sphereface} protocol. Any image that does not contain a face, or has more than one face, is discarded. Finally, for each remaining image, we extract its ArcFace~\cite{deng2019arcface} embedding.

We start by defining the similarity of faces $i$ and $j$ as the cosine distance of their respective embeddings $\textbf{e}_i$, $\textbf{e}_j$:

\begin{equation}
    D_C\left(\textbf{e}_i, \textbf{e}_j\right) = 1 - \frac{\textbf{e}_i \cdot \textbf{e}_j}{{\| \textbf{e}_i \|}_2 {\| \textbf{e}_j \|}_2}
\end{equation}

A cosine distance $D_C$ closer to 0 (or more specifically, under a given threshold $t$) means that the two images' embeddings are similar.
Therefore, if $D_C\left(\textbf{e}_i, \textbf{e}_j\right) \leq t$, we can conclude that $i$ and $j$ have the same facial identity.

The first step is to ensure that images within the same pseudo-class are consistent, \textit{i.e.} all images are similar enough to all other images of the same class. Moreover, we require that each pseudo-class must contain at least one ISO/ICAO-compliant image.

More formally, given a set $I$ of $n$ images and a subset $C \subseteq I$ of ISO/ICAO-compliant images in the same given pseudo-class, we find the largest subset $V \subseteq I$ so that:

\begin{equation}
    \forall \left\{i, j\right\} \in V \quad D_C\left(\textbf{e}_i, \textbf{e}_j\right) \leq t \land V \cap C \neq \emptyset
    \label{eq:intraclass-sim}
\end{equation}

To find the above-mentioned subset, we start by constructing a similarity matrix $S \in \mathbb{R}^{n \times n}$, which is defined as

\begin{equation}
    S_{i,j} =
        \begin{cases}
            1 & \text{if } i \neq j \land D_C\left(\textbf{e}_i, \textbf{e}_j\right) \leq t \\
            0 & \text{otherwise}
        \end{cases}
    \label{eq:similarity-matrix}
\end{equation}

The symmetric binary matrix $S$ can be interpreted as an adjacency matrix of an unweighted undirected similarity graph $G$, where each node represents an identity and each edge indicates that two faces are similar enough.
Then, the set $V$ that satisfies Equation~\ref{eq:intraclass-sim} is found as the largest maximal clique in $G$ that contains at least one ISO/ICAO-compliant image; all other images that are not part of such clique are discarded.
To enumerate all maximal cliques we employ the Bron-Kerbosh algorithm~\cite{bron1973algorithm} with pivoting~\cite{tomita2006worst,cazals2008note}.
Despite having a worst-case time complexity of $\mathcal{O}\left(3^{V/3}\right)$, the running time of the algorithm remains practical given that the graph $G$ contains at most $64$ nodes.

The result of this procedure is shown in Figure~\ref{fig:samples_same_subject}, in which for each line we report the images in the same pseudo-class that we include in the dataset in addition to the ISO/ICAO compliant ones.

\subsection{Inter-class consistency} \label{sec:inter_consistency}
As the prompts for the different pseudo-classes may generate subjects that are too similar to each other, the next step is to select the pseudo-classes that contain faces that are all dissimilar enough.

More formally, given the set of all $n$ pseudo-classes $P$, we want to find a subset of classes $Q \subseteq P$ so that:

\begin{equation}
    \forall \left\{i, j\right\} \in Q, \, i \neq j \quad D_C\left(\textbf{e}_i, \textbf{e}_j\right) > t
\end{equation}

Note that after this step we can refer to the elements of $Q$ as proper classes because each one contains exactly only one homogeneous identity, and different classes represent different identities.
\begin{figure*}[t!]
    \newcommand{\imagewidth}{0.16}
    \centering
        \begin{subfigure}[b]{\imagewidth\linewidth}
        \centering
        \includegraphics[width=\linewidth]{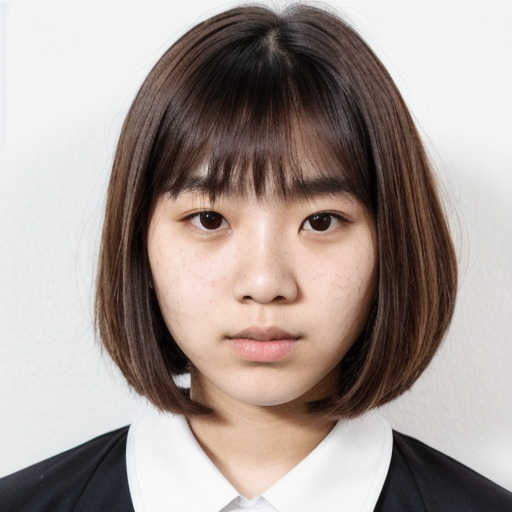}
        \caption{Female}
    \end{subfigure}
    \medspace
    \begin{subfigure}[b]{\imagewidth\linewidth}
        \centering
        \includegraphics[width=\linewidth]{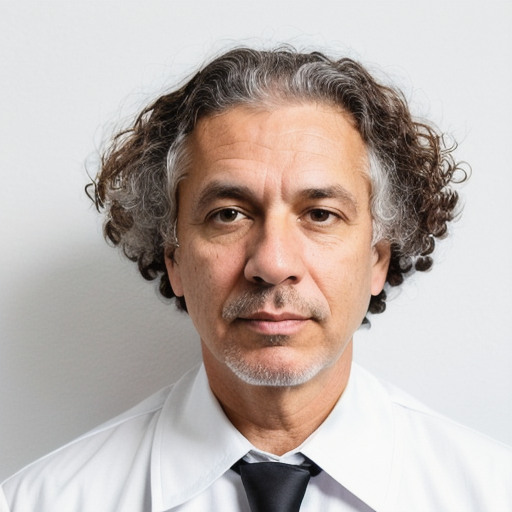}
        \caption{Male}
    \end{subfigure}
    \medspace
    \begin{subfigure}[b]{\imagewidth\linewidth}
        \centering
        \includegraphics[width=\linewidth]{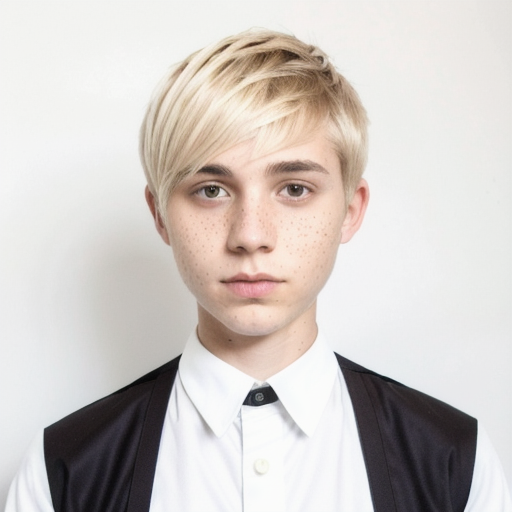}
        \caption{Non-binary}
    \end{subfigure}
    \medspace
    \begin{subfigure}[b]{\imagewidth\linewidth}
        \centering
        \includegraphics[width=\linewidth]{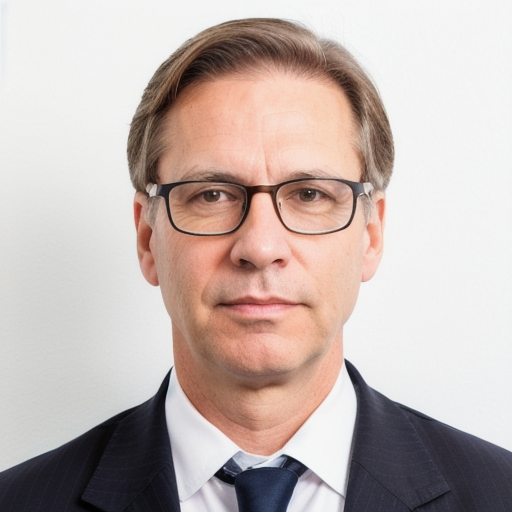}
        \caption{Glasses}
    \end{subfigure}
    \medspace
    \begin{subfigure}[b]{\imagewidth\linewidth}
        \centering
        \includegraphics[width=\linewidth]{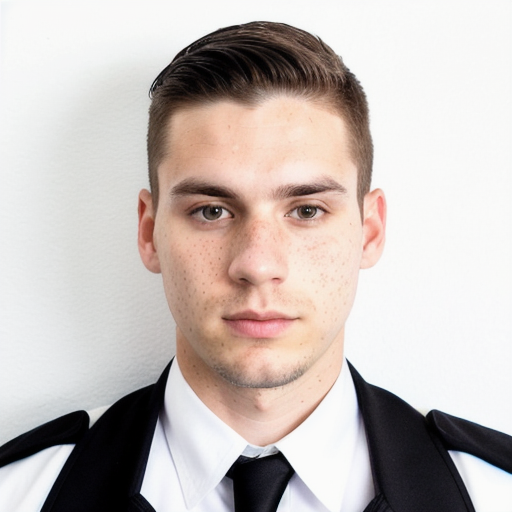}
        \caption{Freckles}
    \end{subfigure}
    \\
    \vspace{0.67em}
    \begin{subfigure}[b]{\imagewidth\linewidth}
        \centering
        \includegraphics[width=\linewidth]{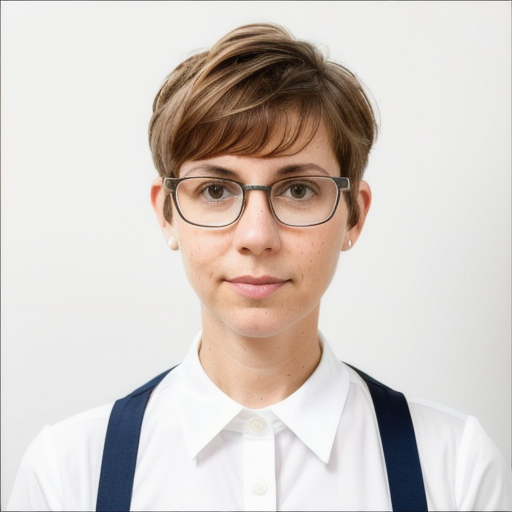}
        \caption{EEA}
    \end{subfigure}
    \medspace
    \begin{subfigure}[b]{\imagewidth\linewidth}
        \centering
        \includegraphics[width=\linewidth]{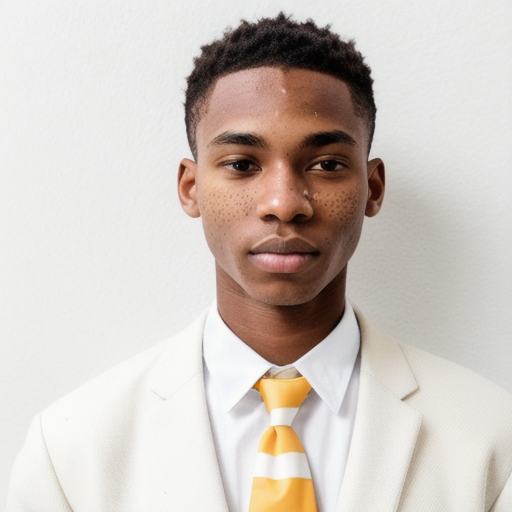}
        \caption{EAF}
    \end{subfigure}
    \medspace
    \begin{subfigure}[b]{\imagewidth\linewidth}
        \centering
        \includegraphics[width=\linewidth]{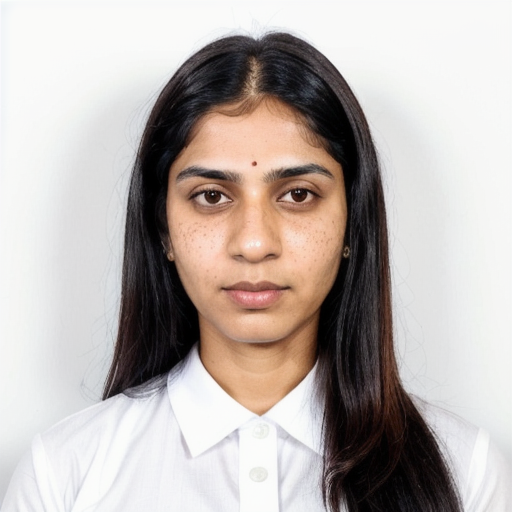}
        \caption{EIA}
    \end{subfigure}
    \medspace
    \begin{subfigure}[b]{\imagewidth\linewidth}
        \centering
        \includegraphics[width=\linewidth]{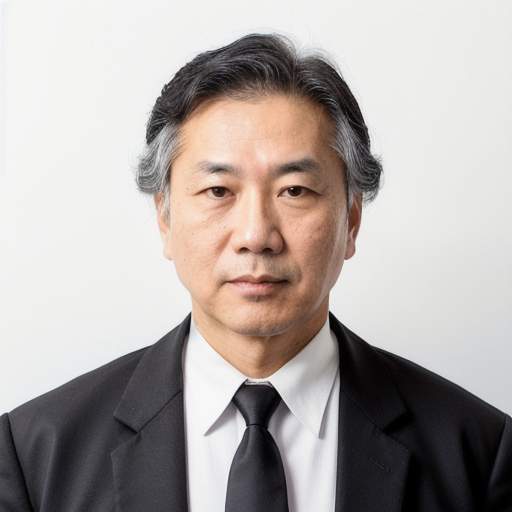}
        \caption{EAS}
    \end{subfigure}
    \medspace
    \begin{subfigure}[b]{\imagewidth\linewidth}
        \centering
        \includegraphics[width=\linewidth]{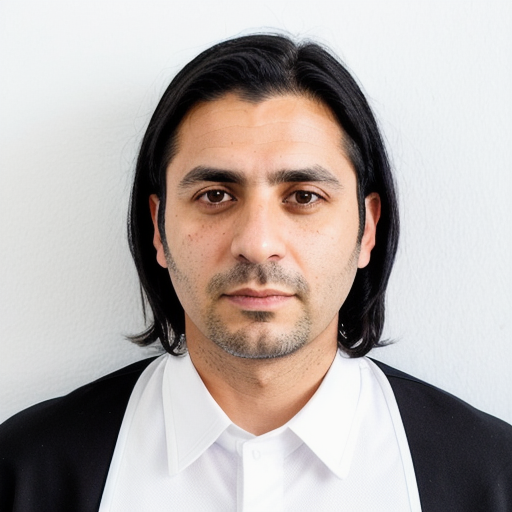}
        \caption{EME}
    \end{subfigure}
    \\
    \vspace{0.67em}
    \begin{subfigure}[b]{\imagewidth\linewidth}
        \centering
        \includegraphics[width=\linewidth]{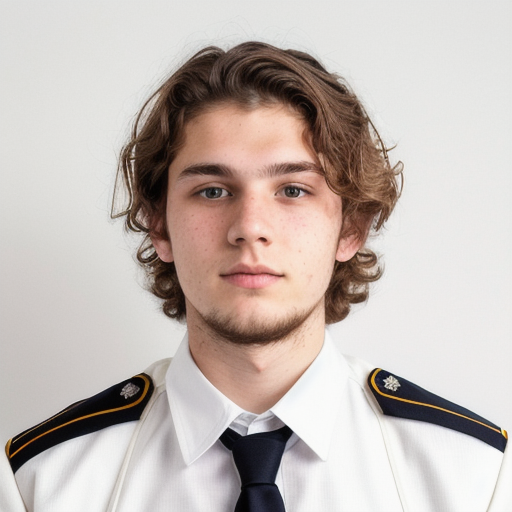}
        \caption{18 y.o.}
    \end{subfigure}
    \medspace
    \begin{subfigure}[b]{\imagewidth\linewidth}
        \centering
        \includegraphics[width=\linewidth]{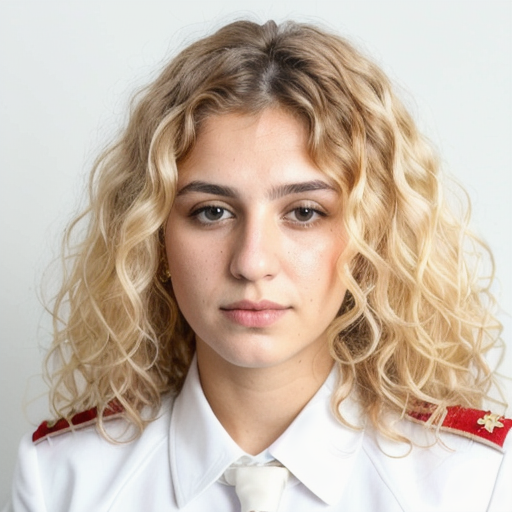}
        \caption{30 y.o.}
    \end{subfigure}
    \medspace
    \begin{subfigure}[b]{\imagewidth\linewidth}
        \centering
        \includegraphics[width=\linewidth]{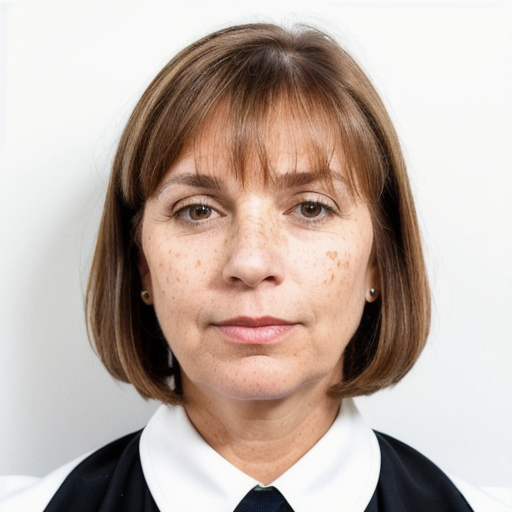}
        \caption{50 y.o.}
    \end{subfigure}
    \medspace
    \begin{subfigure}[b]{\imagewidth\linewidth}
        \centering
        \includegraphics[width=\linewidth]{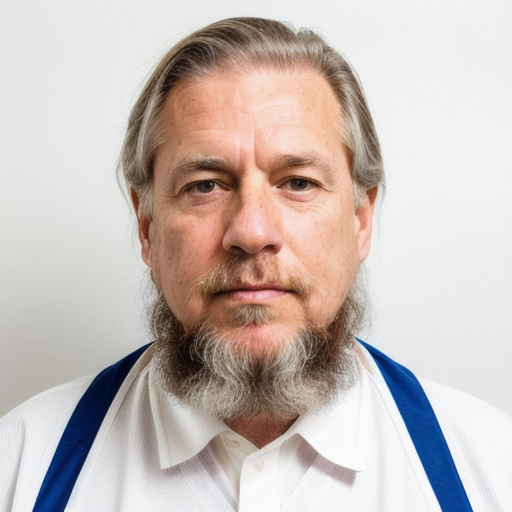}
        \caption{70 y.o.}
    \end{subfigure}
    \medspace
    \begin{subfigure}[b]{\imagewidth\linewidth}
        \centering
        \includegraphics[width=\linewidth]{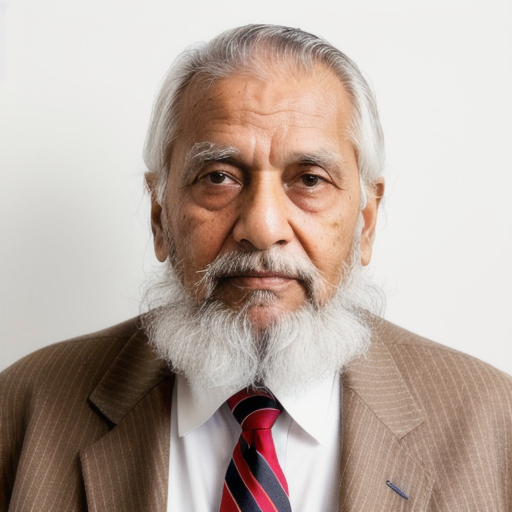}
        \caption{80 y.o.}
    \end{subfigure}
    \caption{Samples of the subject variability included in the ONOT dataset. Different genders, ethnicities, ages and facial traits are included in the dataset, enhancing the variability of the dataset.  The naming convention is reported in Table~\ref{tab:file-naming}.}
    \label{fig:variability}
\end{figure*}
To find $Q$, we compute the similarity matrix $S \in \mathbb{R}^{n \times n}$ as defined by Equation~\ref{eq:similarity-matrix}; each cell represents the comparison of the ISO/ICAO-compliant images' embeddings across all pseudo-classes.
As before, $S$ can be interpreted as an adjacency matrix of an unweighted undirected similarity graph $G$, where each node represents a pseudo-class and each edge indicates that two pseudo-classes are similar enough.
Then, we note that finding $Q$ consists of computing the maximum independent set in $G$. All pseudo-classes not part of the found subset of nodes are discarded.

\subsection{ICAO and identity consistency test statistics}\label{sec:thresholds}
As mentioned, the initial dataset generation includes $15$k different pseudo-classes. After the first ISO/ICAO compliance test, $4032$ identities survive: this reduction ($-73$\%) indicates a certain complexity in controlling specific face characteristics, as further analyzed in the next section. 

The following inter- and intra-class consistency tests are based on a given threshold ($t$) for the face verification system~\cite{deng2019arcface}: in our case, we employ three distinct thresholds, experimentally determined by the execution of a set of $20$k impostor face verification attempts on a separate real face dataset. 
The identified thresholds correspond to the $\text{FMR}_{100}$, $\text{FMR}_{1000}$, $\text{FMR}_{10000}$, and are respectively $0.597$, $0.493$, and $0.413$.
We obtained this way three image subsets; note that as the threshold values increase, the number of images within each class grows, while the count of distinct classes decreases.
Specifically, after the identity consistency test $55$, $125$ and $255$ distinct identities remain, for the three thresholds, respectively. 
These numbers reveal the challenges of generating faces that combine strict ICAO-compliant requirements and identity-based checks.

\begin{figure}[t!]
    \newcommand{\imagewidth}{0.32}
    \centering
    \begin{subfigure}[b]{\imagewidth\linewidth}
        \centering
        \includegraphics[width=\linewidth]{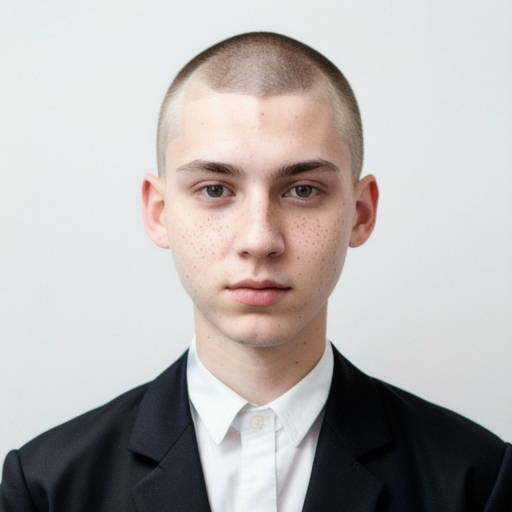}
    \end{subfigure}
    \medspace
    \begin{subfigure}[b]{\imagewidth\linewidth}
        \centering
        \includegraphics[width=\linewidth]{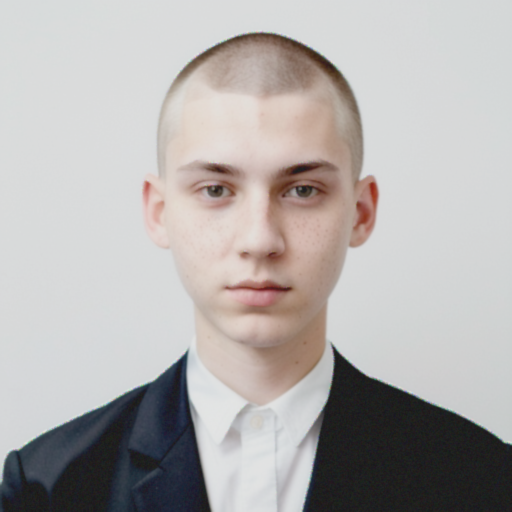}
    \end{subfigure}
    \\
    \vspace{0.67em}
    \medspace
    \begin{subfigure}[b]{\imagewidth\linewidth}
        \centering
        \includegraphics[width=\linewidth]{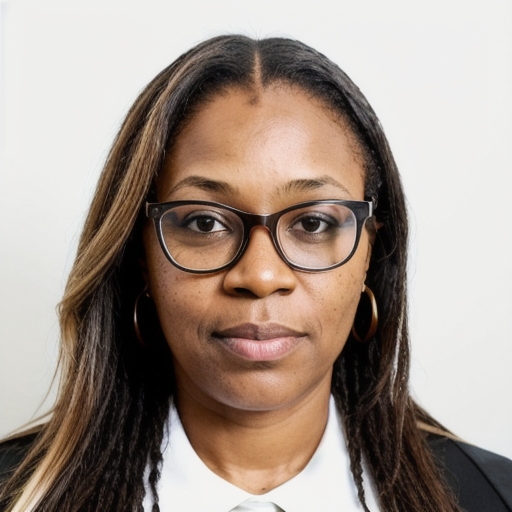}
        \caption{Digital}
    \end{subfigure}
    \medspace
    \begin{subfigure}[b]{\imagewidth\linewidth}
        \centering
        \includegraphics[width=\linewidth]{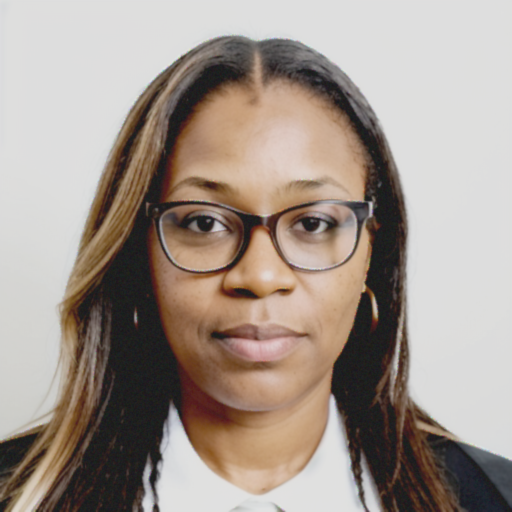}
        \caption{P\&S}
    \end{subfigure}
    \caption{Visual samples of the application of the P\&S operation (see Sect.~\ref{sec:pes}) on two original images.}
    \label{fig:pes}
\end{figure}

\begin{figure*}[th!]
    \centering
    \includegraphics[width=0.9\linewidth]{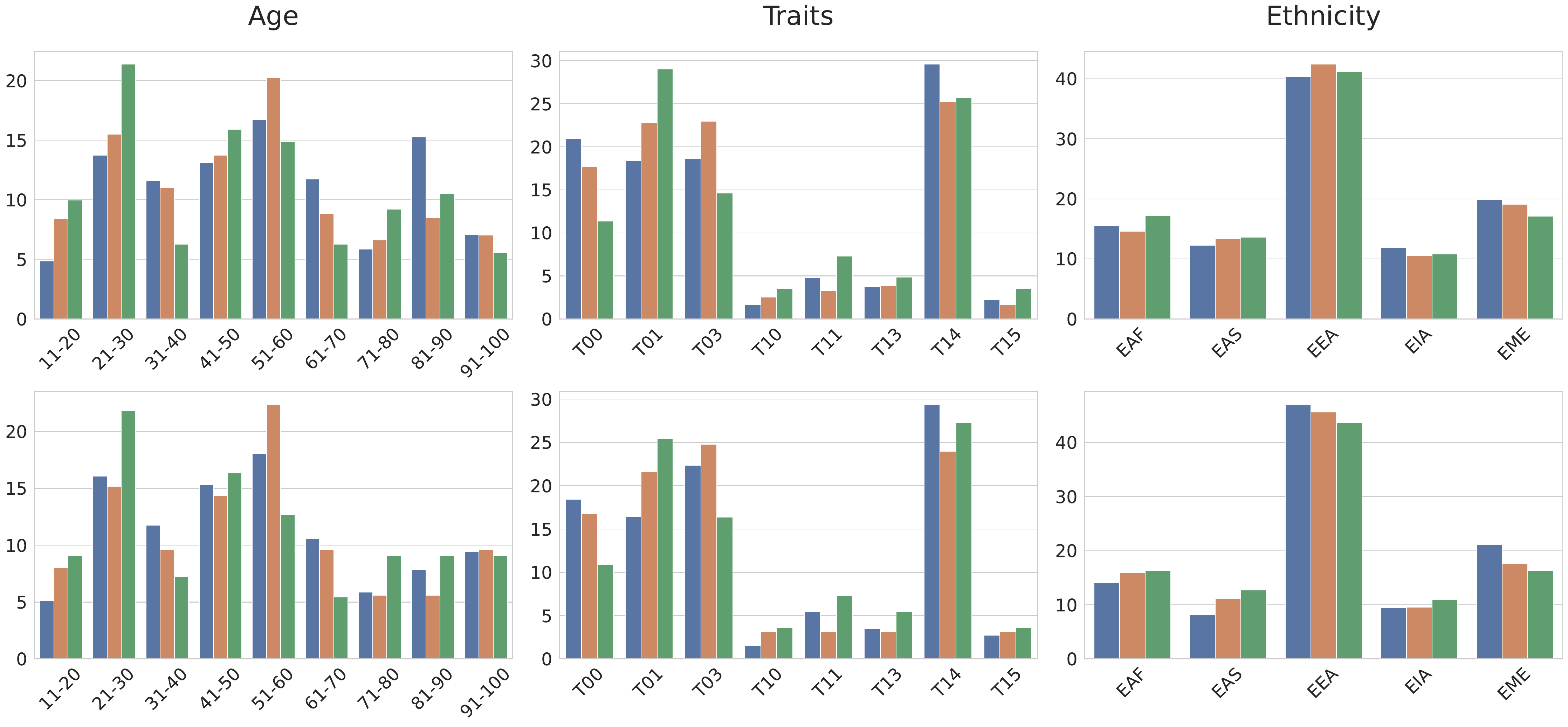}
    \caption{Age, traits, and ethnicity distributions of ONOT images. The blue, orange, and green bars respectively denote images belonging to subsets defined by the three thresholds defined in Section~\ref{sec:thresholds}. The bottom three plots include only ISO/ICAO-compliant images, while the top three include all images. The naming convention is reported in Table~\ref{tab:file-naming}.
    }
    \label{fig:stats-plots}
\end{figure*}

We observe these distinct subsets correspond to different working scenarios: for instance, the use of a low threshold implies a high level of similarity across different identities and a lower intra-class variability, representing a challenging benchmark for face analysis tasks since the resulting dataset will include several cases of look-alike subjects. 
Vice versa, a higher threshold implies the presence of more distinct identities, but a higher level of intra-class variability, making it suitable, for instance, to improve the robustness of FRSs to typical variations of face appearance.

The ONOT dataset is released including the index annotation files needed to reproduce the three subsets.

\subsection{Print\&Scan Generation} \label{sec:pes}
Finally, for each image available in the ONOT dataset, we simulate the print and scan process (P\&S) through the method described in \cite{ferrara2021face}.
We include the P\&S operation since it is typical in procedures related to the issuance of electronic identity documents~\cite{borghi2023revelio}: such processes commonly entail the submission of a passport-sized photograph, which is later scanned and compressed for storage in the document's chip. P\&S images are released together with the original version of the ONOT dataset. Figure \ref{fig:pes} provides a visual representation of the results of these operations.

\section{DATASET STATISTICS}

%



Some examples representing the variability of the dataset are depicted in Figure~\ref{fig:variability}.
In addition, main dataset statistics are illustrated in Figure~\ref{fig:stats-plots}, in which the first line shows plots computed on all the dataset images, while the second row the plots computed only on the ISO/ICAO compliant images.



\begin{figure*}
    \centering
    \includegraphics[width=0.91\linewidth]{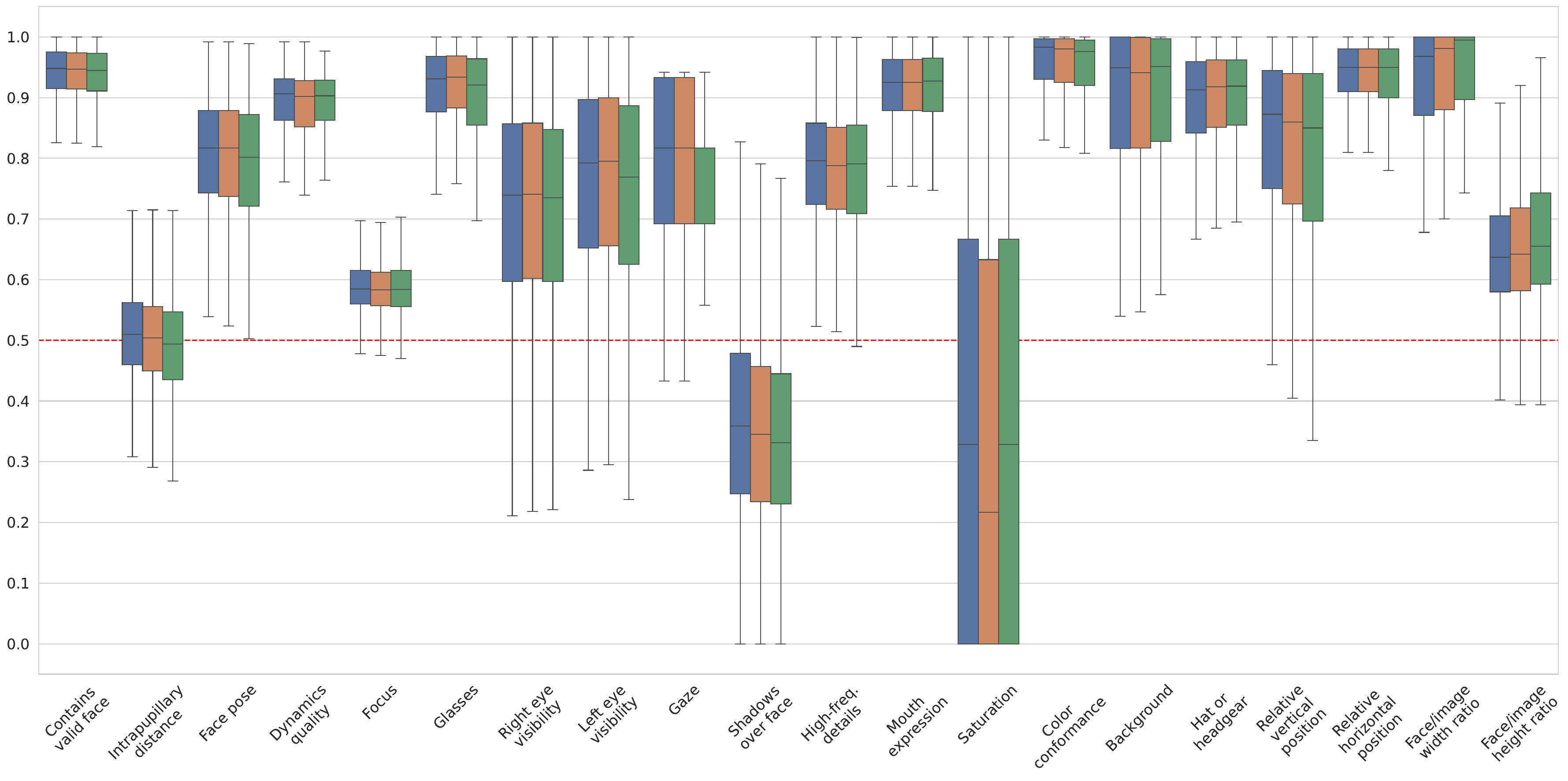}
    \caption{Scores distribution of the tests performed during ISO/ICAO compliance check.
    An image is considered compliant if all checks return a score greater than $0.5$ (dashed red line). The blue, orange, and green boxes respectively denote images belonging to subsets defined in Section \ref{sec:inter_consistency}. Each box represents the quartiles of each score distribution, while the whiskers encompass the whole values' range. 
    }
    \label{fig:stats-dist-all}
\end{figure*}

We observe that the majority of the identities that pass the ICAO and identity consistency tests are in the range of $[21-60]$ years old, with the main peaks located in $[21-30]$ and $[51-60]$.
The second plots reveal that the most common trait is represented by the combination of freckles and moles (T14), followed by no specific attributes (T00), freckles (T01) and moles (T03): these percentages follow the distribution detailed at the beginning of this section. 
Thus, this indicates that generating faces with specific attributes does not significantly influence compliance with the ICAO test.
Noticeably, the third plots reveal the presence of a significant ethnic bias toward caucasian (EEA) subjects, which comprises more than $40$\% of the dataset regardless of the chosen subset.
%
%
To investigate this behavior, we evaluate the proportion of images grouped by ethnicity relative to the total number of images before and after the checks detailed in Section~\ref{sec:iso_check}.
In particular, the proportion of images that depict a caucasian subject initially accounts for $20.4$\% of the dataset (since $5$ different ethnicities are considered). This proportion significantly increases to $45.6$\% after the ICAO and identity consistency tests.
Conversely, the East-Asian (EAS) and Indian-Asian (EIA) ethnicities experience a substantial reduction in representation, decreasing from $20.4$\% and $19.8$\% to $6.0$\% and $7.1$\% respectively.
Finally, Middle Eastern (EME) and African (EAF) ethnicities exhibit minimal variation in representation in the dataset.

These observations indicate a potential bias in the employed face verification, which is less able to discriminate identities in no-caucasian ethnicities, as suggested also in the literature~\cite{cavazos2020accuracy}.
With respect to the commercial ICAO SDK exploited, the slight difference in the distribution of the two rows denotes that the software is more robust, having a more uniform behavior across all ethnicities.  
Besides, these values can also indicate a complexity in the generation of images of a specific ethnicity, due to, for instance, an underrepresentation bias in data used for training the generative model.  


Moreover, we plot the scores' distribution for the tests reported in Table~\ref{tab:test_iso} in Figure~\ref{fig:stats-dist-all}.
These scores are output by the commercial ICAO SDK validation tool and each produced score is in the range $[0, 1]$. 
Following the official guideline of the SDK, an image is considered ISO/ICAO-compliant if all tests return a score greater or equal to $0.5$ (red line in the plot).
Results indicate that two tests, specifically ``shadows over face'' and ``saturation'', pose notable challenges for the images in the dataset. This highlights the difficulty faced by the image generator in controlling lighting conditions and saturation of the generated images, therefore significantly impacting the number of images that pass the ISO/ICAO compliance checks.

\begin{figure}[th!]
    \newcommand{\imagewidth}{0.23}
    \centering
    \begin{subfigure}[b]{\imagewidth\linewidth}
        \centering
        \includegraphics[width=\linewidth]{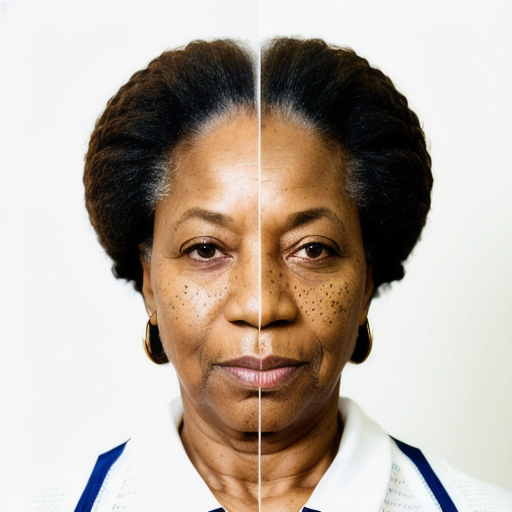}
    \end{subfigure}
    \medspace
    \begin{subfigure}[b]{\imagewidth\linewidth}
        \centering
        \includegraphics[width=\linewidth]{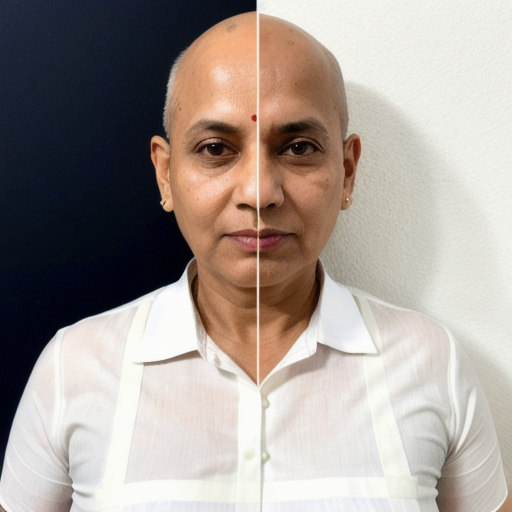}
    \end{subfigure}
    \medspace
    \begin{subfigure}[b]{\imagewidth\linewidth}
        \centering
        \includegraphics[width=\linewidth]{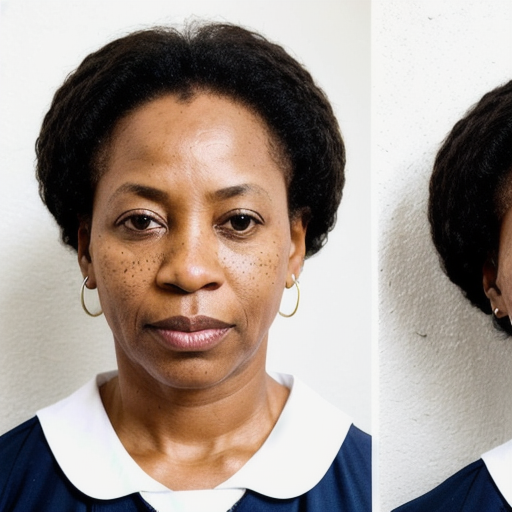}
    \end{subfigure}
    \medspace
    \begin{subfigure}[b]{\imagewidth\linewidth}
        \centering
        \includegraphics[width=\linewidth]{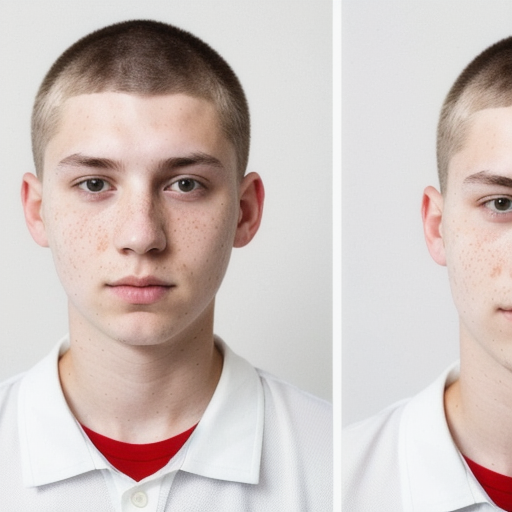}
    \end{subfigure}
    \caption{Failure cases of the generation procedure.}
    \label{fig:failures}
\end{figure}

\section{DISCUSSION AND FUTURE WORK}
Despite the high quality of the generated images, the prompt-based generation is complex, especially when strict quality requirements have to be fulfilled. Indeed, there are specific facial characteristics particularly challenging to accurately control during the generation such as, as previously mentioned, uniform lighting and skin color and the presence of shadows. We hypothesize this is due to the nature of the images used for training the generative model, mainly belonging to unconstrained real-world scenarios with images including, for instance, flashes and glares. 
Furthermore, we observe the presence of a limited number of images presenting some generation artifacts, depicted in Figure~\ref{fig:failures}, that are difficult to detect through automatic face verification or face quality analysis systems.
For instance, the first two images are correctly detected as faces by the used face detector~\cite{zhang2016joint}, while in the remaining images only a single face is detected.

Another critical aspect is the variability in generated identities: indeed, starting from $15$k pseudo-classes, only $255$ identities survive the selection procedures (ICAO compliance and identity consistency tests).
In consideration of this difficulty in controlling generated identities, we replicated a similar generation experiment by forcing in the positive prompt the generation of specific identities associated with well-known individuals (\textit{e.g.} actors, politicians, and the like). In this case, starting from the same number of initial pseudo-classes, there was an observed increase of $37$\% in surviving identities, denoting the complexity of generating anonymous identities from seed with respect to the generation of images of known identities.
Moreover, we observe a certain complexity also in generating multiple images of the same identity: employing the same prompt with slightly different seeds does not guarantee a constant identity across the generated images, thus requiring intra-class consistency tests. 

Finally, another significant challenge is posed by the ISO/ICAO compliance verification tool: as many of these tools are commercial and closed-source, the precise reasons for a particular image failing a specific quality test are difficult to determine; therefore, engineering prompts that maximize the number of images that pass the ISO/ICAO compliance checks proves to be arduous.

As future work, we plan to increment the number of generated identities, and the release of a novel set of morphed images created starting from ONOT bona fide subjects, using multiple morphing algorithms. Another research topic should regard the possibility of constraining the generation through not only the input prompt, but also exploiting additional multi-modal information sources (\textit{e.g.} models, images) that control specific elements (\textit{e.g.} head pose, identity, age).

\section*{ACKNOWLEDGMENTS}
This project received funding from the European Union’s Horizon 2020 research and
innovation program under Grant Agreement No. 883356. Disclaimer: this text reflects only the
author’s views, and the Commission is not liable for any use that may be made of the information
contained therein.

We thank Andrea Pilzer, NVIDIA AI Technology Center, EMEA, for his support. We also acknowledge the CINECA award under the ISCRA initiative, for the availability of high-performance computing resources and support.


{\small
\bibliographystyle{ieee}
\bibliography{main}

\begin{thebibliography}{10}\itemsep=-1pt

\bibitem{abay2019privacy}
N.~C. Abay, Y.~Zhou, M.~Kantarcioglu, B.~Thuraisingham, and L.~Sweeney.
\newblock Privacy preserving synthetic data release using deep learning.
\newblock In {\em Machine Learning and Knowledge Discovery in Databases: European Conference, ECML PKDD 2018, Dublin, Ireland, September 10--14, 2018, Proceedings, Part I 18}, pages 510--526. Springer, 2019.

\bibitem{AUTOMATIC1111_Stable_Diffusion_Web_2022}
AUTOMATIC1111.
\newblock {Stable Diffusion Web UI}, 2022.

\bibitem{bae2023digiface}
G.~Bae, M.~de~La~Gorce, T.~Baltru{\v{s}}aitis, C.~Hewitt, D.~Chen, J.~Valentin, R.~Cipolla, and J.~Shen.
\newblock Digiface-1m: 1 million digital face images for face recognition.
\newblock In {\em Proceedings of the IEEE/CVF Winter Conference on Applications of Computer Vision}, pages 3526--3535, 2023.

\bibitem{borghi2023revelio}
G.~Borghi, N.~Di~Domenico, A.~Franco, M.~Ferrara, and D.~Maltoni.
\newblock Revelio: A modular and effective framework for reproducible training and evaluation of morphing attack detectors.
\newblock {\em IEEE Access}, 2023.

\bibitem{borghi2021double}
G.~Borghi, E.~Pancisi, M.~Ferrara, and D.~Maltoni.
\newblock A double siamese framework for differential morphing attack detection.
\newblock {\em Sensors}, 21(10):3466, 2021.

\bibitem{boutros2023idiff}
F.~Boutros, J.~H. Grebe, A.~Kuijper, and N.~Damer.
\newblock Idiff-face: Synthetic-based face recognition through fizzy identity-conditioned diffusion model.
\newblock In {\em Proceedings of the IEEE/CVF International Conference on Computer Vision}, pages 19650--19661, 2023.

\bibitem{boutros2022sface}
F.~Boutros, M.~Huber, P.~Siebke, T.~Rieber, and N.~Damer.
\newblock Sface: Privacy-friendly and accurate face recognition using synthetic data.
\newblock In {\em 2022 IEEE International Joint Conference on Biometrics (IJCB)}, pages 1--11. IEEE, 2022.

\bibitem{boutros2023unsupervised}
F.~Boutros, M.~Klemt, M.~Fang, A.~Kuijper, and N.~Damer.
\newblock Unsupervised face recognition using unlabeled synthetic data.
\newblock In {\em 2023 IEEE 17th International Conference on Automatic Face and Gesture Recognition (FG)}, pages 1--8. IEEE, 2023.

\bibitem{boutros2023synthetic}
F.~Boutros, V.~Struc, J.~Fierrez, and N.~Damer.
\newblock Synthetic data for face recognition: Current state and future prospects.
\newblock {\em Image and Vision Computing}, page 104688, 2023.

\bibitem{bowyer2004face}
K.~W. Bowyer.
\newblock Face recognition technology: security versus privacy.
\newblock {\em IEEE Technology and society magazine}, 23(1):9--19, 2004.

\bibitem{bron1973algorithm}
C.~Bron and J.~Kerbosch.
\newblock Algorithm 457: finding all cliques of an undirected graph.
\newblock {\em Communications of the ACM}, 16(9):575--577, 1973.

\bibitem{busch2023standards}
C.~Busch.
\newblock Standards for biometric presentation attack detection.
\newblock In {\em Handbook of Biometric Anti-Spoofing: Presentation Attack Detection and Vulnerability Assessment}, pages 571--583. Springer, 2023.

\bibitem{cavazos2020accuracy}
J.~G. Cavazos, P.~J. Phillips, C.~D. Castillo, and A.~J. O’Toole.
\newblock Accuracy comparison across face recognition algorithms: Where are we on measuring race bias?
\newblock {\em IEEE transactions on biometrics, behavior, and identity science}, 3(1):101--111, 2020.

\bibitem{cazals2008note}
F.~Cazals and C.~Karande.
\newblock A note on the problem of reporting maximal cliques.
\newblock {\em Theoretical computer science}, 407(1-3):564--568, 2008.

\bibitem{damer2022privacy}
N.~Damer, C.~A.~F. L{\'o}pez, M.~Fang, N.~Spiller, M.~V. Pham, and F.~Boutros.
\newblock Privacy-friendly synthetic data for the development of face morphing attack detectors.
\newblock In {\em Proceedings of the IEEE/CVF Conference on Computer Vision and Pattern Recognition}, pages 1606--1617, 2022.

\bibitem{deng2019arcface}
J.~Deng, J.~Guo, N.~Xue, and S.~Zafeiriou.
\newblock Arcface: Additive angular margin loss for deep face recognition.
\newblock In {\em Proceedings of the IEEE/CVF conference on computer vision and pattern recognition}, pages 4690--4699, 2019.

\bibitem{deng2020disentangled}
Y.~Deng, J.~Yang, D.~Chen, F.~Wen, and X.~Tong.
\newblock Disentangled and controllable face image generation via 3d imitative-contrastive learning.
\newblock In {\em Proceedings of the IEEE/CVF conference on computer vision and pattern recognition}, pages 5154--5163, 2020.

\bibitem{dhariwal2021diffusion}
P.~Dhariwal and A.~Nichol.
\newblock Diffusion models beat gans on image synthesis.
\newblock {\em Advances in neural information processing systems}, 34:8780--8794, 2021.

\bibitem{ferrara2021face}
M.~Ferrara, A.~Franco, and D.~Maltoni.
\newblock Face morphing detection in the presence of printing/scanning and heterogeneous image sources.
\newblock {\em IET Biometrics}, 10(3):290--303, 2021.

\bibitem{franco2022face}
A.~Franco, A.~Magnani, D.~Maltoni, D.~Maio, L.~Odorisio, and A.~De~Maria.
\newblock Face image quality assessment in electronic id documents.
\newblock {\em IEEE Access}, 10:77744--77758, 2022.

\bibitem{frigieri2017fast}
E.~Frigieri, G.~Borghi, R.~Vezzani, and R.~Cucchiara.
\newblock Fast and accurate facial landmark localization in depth images for in-car applications.
\newblock In {\em Image Analysis and Processing-ICIAP 2017: 19th International Conference, Catania, Italy, September 11-15, 2017, Proceedings, Part I 19}, pages 539--549. Springer, 2017.

\bibitem{goodfellow2020generative}
I.~Goodfellow, J.~Pouget-Abadie, M.~Mirza, B.~Xu, D.~Warde-Farley, S.~Ozair, A.~Courville, and Y.~Bengio.
\newblock Generative adversarial networks.
\newblock {\em Communications of the ACM}, 63(11):139--144, 2020.

\bibitem{ISO-19794-5}
{ISO/IEC 19794-5 — Information technology — Biometric data interchange formats — Part 5: Face image data}.
\newblock Standard, International Organization for Standardization, 2011.

\bibitem{ISO}
{ISO/IEC 39794-5 — Information technology — Extensible biometric data interchange formats — Part 5: Face image data}.
\newblock Standard, International Organization for Standardization, 2019.

\bibitem{kansy2023controllable}
M.~Kansy, A.~Ra{\"e}l, G.~Mignone, J.~Naruniec, C.~Schroers, M.~Gross, and R.~M. Weber.
\newblock Controllable inversion of black-box face recognition models via diffusion.
\newblock In {\em Proceedings of the IEEE/CVF International Conference on Computer Vision}, pages 3167--3177, 2023.

\bibitem{karkkainen2021fairface}
K.~Karkkainen and J.~Joo.
\newblock Fairface: Face attribute dataset for balanced race, gender, and age for bias measurement and mitigation.
\newblock In {\em Proceedings of the IEEE/CVF winter conference on applications of computer vision}, pages 1548--1558, 2021.

\bibitem{karras2022elucidating}
T.~Karras, M.~Aittala, T.~Aila, and S.~Laine.
\newblock Elucidating the design space of diffusion-based generative models.
\newblock {\em Advances in Neural Information Processing Systems}, 35:26565--26577, 2022.

\bibitem{karras2020training}
T.~Karras, M.~Aittala, J.~Hellsten, S.~Laine, J.~Lehtinen, and T.~Aila.
\newblock Training generative adversarial networks with limited data.
\newblock {\em Advances in neural information processing systems}, 33:12104--12114, 2020.

\bibitem{kingma2019introduction}
D.~P. Kingma, M.~Welling, et~al.
\newblock An introduction to variational autoencoders.
\newblock {\em Foundations and Trends{\textregistered} in Machine Learning}, 12(4):307--392, 2019.

\bibitem{li2020deep}
S.~Li and W.~Deng.
\newblock Deep facial expression recognition: A survey.
\newblock {\em IEEE transactions on affective computing}, 13(3):1195--1215, 2020.

\bibitem{liu2017sphereface}
W.~Liu, Y.~Wen, Z.~Yu, M.~Li, B.~Raj, and L.~Song.
\newblock Sphereface: Deep hypersphere embedding for face recognition.
\newblock In {\em Proceedings of the IEEE conference on computer vision and pattern recognition}, pages 212--220, 2017.

\bibitem{lu2022dpm}
C.~Lu, Y.~Zhou, F.~Bao, J.~Chen, C.~Li, and J.~Zhu.
\newblock Dpm-solver++: Fast solver for guided sampling of diffusion probabilistic models.
\newblock {\em arXiv preprint arXiv:2211.01095}, 2022.

\bibitem{melzi2023gandiffface}
P.~Melzi, C.~Rathgeb, R.~Tolosana, R.~Vera-Rodriguez, D.~Lawatsch, F.~Domin, and M.~Schaubert.
\newblock Gandiffface: Controllable generation of synthetic datasets for face recognition with realistic variations.
\newblock {\em arXiv preprint arXiv:2305.19962}, 2023.

\bibitem{melzi2024frcsyn}
P.~Melzi, R.~Tolosana, R.~Vera-Rodriguez, M.~Kim, C.~Rathgeb, X.~Liu, I.~DeAndres-Tame, A.~Morales, J.~Fierrez, J.~Ortega-Garcia, et~al.
\newblock Frcsyn-ongoing: Benchmarking and comprehensive evaluation of real and synthetic data to improve face recognition systems.
\newblock {\em Information Fusion}, page 102322, 2024.

\bibitem{nichol2021improved}
A.~Q. Nichol and P.~Dhariwal.
\newblock Improved denoising diffusion probabilistic models.
\newblock In {\em International Conference on Machine Learning}, pages 8162--8171. PMLR, 2021.

\bibitem{ntoutsi2020bias}
E.~Ntoutsi, P.~Fafalios, U.~Gadiraju, V.~Iosifidis, W.~Nejdl, M.-E. Vidal, S.~Ruggieri, F.~Turini, S.~Papadopoulos, E.~Krasanakis, et~al.
\newblock Bias in data-driven artificial intelligence systems—an introductory survey.
\newblock {\em Wiley Interdisciplinary Reviews: Data Mining and Knowledge Discovery}, 10(3):e1356, 2020.

\bibitem{qiu2021synface}
H.~Qiu, B.~Yu, D.~Gong, Z.~Li, W.~Liu, and D.~Tao.
\newblock Synface: Face recognition with synthetic data.
\newblock In {\em Proceedings of the IEEE/CVF International Conference on Computer Vision}, pages 10880--10890, 2021.

\bibitem{raja2020morphing}
K.~Raja, M.~Ferrara, A.~Franco, L.~Spreeuwers, I.~Batskos, F.~de~Wit, M.~Gomez-Barrero, U.~Scherhag, D.~Fischer, S.~K. Venkatesh, et~al.
\newblock Morphing attack detection-database, evaluation platform, and benchmarking.
\newblock {\em IEEE transactions on information forensics and security}, 16:4336--4351, 2020.

\bibitem{Rombach2022CVPR}
R.~Rombach, A.~Blattmann, D.~Lorenz, P.~Esser, and B.~Ommer.
\newblock High-resolution image synthesis with latent diffusion models.
\newblock In {\em Proceedings of the IEEE/CVF Conference on Computer Vision and Pattern Recognition (CVPR)}, pages 10684--10695, June 2022.

\bibitem{schlett2022face}
T.~Schlett, C.~Rathgeb, O.~Henniger, J.~Galbally, J.~Fierrez, and C.~Busch.
\newblock Face image quality assessment: A literature survey.
\newblock {\em ACM Computing Surveys (CSUR)}, 54(10s):1--49, 2022.

\bibitem{schroff2015facenet}
F.~Schroff, D.~Kalenichenko, and J.~Philbin.
\newblock Facenet: A unified embedding for face recognition and clustering.
\newblock In {\em Proceedings of the IEEE conference on computer vision and pattern recognition}, pages 815--823, 2015.

\bibitem{taigman2015web}
Y.~Taigman, M.~Yang, M.~Ranzato, and L.~Wolf.
\newblock Web-scale training for face identification.
\newblock In {\em Proceedings of the IEEE conference on computer vision and pattern recognition}, pages 2746--2754, 2015.

\bibitem{tian2011facial}
Y.~Tian, T.~Kanade, and J.~F. Cohn.
\newblock Facial expression recognition.
\newblock {\em Handbook of face recognition}, pages 487--519, 2011.

\bibitem{tomita2006worst}
E.~Tomita, A.~Tanaka, and H.~Takahashi.
\newblock The worst-case time complexity for generating all maximal cliques and computational experiments.
\newblock {\em Theoretical computer science}, 363(1):28--42, 2006.

\bibitem{venkatesh2021face}
S.~Venkatesh, R.~Ramachandra, K.~Raja, and C.~Busch.
\newblock Face morphing attack generation and detection: A comprehensive survey.
\newblock {\em IEEE transactions on technology and society}, 2(3):128--145, 2021.

\bibitem{wolf2018icao}
A.~Wolf.
\newblock {ICAO}: Portrait quality (reference facial images for {MRTD}), version 1.0. standard.
\newblock {\em International Civil Aviation Organization}, 2018.

\bibitem{wood2021fake}
E.~Wood, T.~Baltru{\v{s}}aitis, C.~Hewitt, S.~Dziadzio, T.~J. Cashman, and J.~Shotton.
\newblock Fake it till you make it: face analysis in the wild using synthetic data alone.
\newblock In {\em Proceedings of the IEEE/CVF international conference on computer vision}, pages 3681--3691, 2021.

\bibitem{wu2019facial}
Y.~Wu and Q.~Ji.
\newblock Facial landmark detection: A literature survey.
\newblock {\em International Journal of Computer Vision}, 127:115--142, 2019.

\bibitem{zhang2016joint}
K.~Zhang, Z.~Zhang, Z.~Li, and Y.~Qiao.
\newblock Joint face detection and alignment using multitask cascaded convolutional networks.
\newblock {\em IEEE signal processing letters}, 23(10):1499--1503, 2016.

\bibitem{zhao2003face}
W.~Zhao, R.~Chellappa, P.~J. Phillips, and A.~Rosenfeld.
\newblock Face recognition: A literature survey.
\newblock {\em ACM computing surveys (CSUR)}, 35(4):399--458, 2003.

\bibitem{zhu2021webface260m}
Z.~Zhu, G.~Huang, J.~Deng, Y.~Ye, J.~Huang, X.~Chen, J.~Zhu, T.~Yang, J.~Lu, D.~Du, et~al.
\newblock Webface260m: A benchmark unveiling the power of million-scale deep face recognition.
\newblock In {\em Proceedings of the IEEE/CVF Conference on Computer Vision and Pattern Recognition}, pages 10492--10502, 2021.

\end{thebibliography}
}

\end{document}